\documentclass{article}
\usepackage{ijcai22}

\usepackage[utf8]{inputenc}
\usepackage[T1]{fontenc}    
\usepackage{amsfonts}       
\usepackage{nicefrac}       
\usepackage{microtype}      
\usepackage{xcolor}         
\usepackage{natbib}
\setcitestyle{open={[},authoryear,close={]}} 

\usepackage{url, hyperref}
\usepackage{algorithm, algpseudocode} %
\renewcommand{\algorithmicrequire}{\textbf{Input:}}
\renewcommand{\algorithmicensure}{\textbf{Output:}}
\usepackage{amsthm, amsmath, amssymb, amsthm} 
\usepackage{enumerate, paralist}
\usepackage{natbib}
\usepackage{comment}
\usepackage{multirow}
\usepackage{xspace}
\usepackage{rotating}
\usepackage{graphicx}

\newtheorem{theorem}{Theorem}
\newtheorem{prove}{Prove}[section]

\newtheorem{definition}{Definition}[section]
\newtheorem{lemma}{Lemma}[section]

\newcommand{\sysn}{\text{CNN-UCB}\xspace }

\newcommand{\ngxt}{g_{\text{cnn}}(\mathbf{x}_t; \btheta_t)}
\newcommand{\ngx}[3][2]{g_{\text{cnn}}(\mathbf{x}_{#3}; \btheta_{#2})}

\newcommand{\fx}{f(\mathbf{x}_t; \btheta_t)}
\newcommand{\hx}{f^\ast(\mathbf{x}_t)}
\newcommand{\deter}{\text{\normalfont det}}

\newcommand{\bw}{\mathbf{W}}

\newcommand{\ucb}{\text{UCB}}

\newcommand{\btheta}{\boldsymbol \theta}
\newcommand{\bx}{\mathbf{x}}
\newcommand{\bxst}{ \{\bx_i\}_{i=1}^t} 
\newcommand{\bbr}{\mathbb{R}}
\newcommand{\bts}{\btheta^{\ast}_t}

\newcommand{\bg}{\mathbf{G}}

\newcommand{\bbe}[1]{\mathbb{E} \left[ #1 \right] }
\newcommand{\bbp}[1]{\mathbb{P} \left [  #1 \right] }
\newcommand{\bba}[1]{\left|  #1 \right|}
\newcommand{\bbd}{\mathbf{D}}
\newcommand{\para}[1]{\textbf{#1}\xspace}

\newcommand{\bwl}{\bw^l}
\newcommand{\gn}{\mathcal{N}}
\newcommand{\interm}{ \left (\bwl \phi(h^{l-1}) \right)}
\newcommand{\bbi}{\mathbf{I}}

\newcommand{\pphi}[1]{\phi \left( #1 \right) }
\newcommand{\dotp}[1]{ \left \langle #1  \right \rangle}

\newcommand{\bbk}{\mathbf{K}}
\newcommand{\ch}{\widetilde{h}}
\newcommand{\oo}[1]{\mathcal{O} \left( #1   \right)}

\newcommand{\bsg}{\mathbf{G}}
\newcommand{\bsf}{\mathbf{f}}
\newcommand{\bsr}{\mathbf{r}}
\newcommand{\wbtheta}{\widetilde{\btheta}}
\newcommand{\whtt}{\widehat{\btheta}_t}

\title{Convolutional Neural Bandit for Visual-aware Recommendation}

\begin{document}

\author{Yikun Ban, Jingrui He \\
University of Illinois Urbana-Champaign    }

\maketitle

\begin{abstract}
Online recommendation/advertising is ubiquitous in web business. Image displaying is considered as one of the most commonly used formats to interact with customers. Contextual multi-armed bandit has shown success in the application of advertising to solve the exploration-exploitation dilemma existing in the recommendation procedure. Inspired by the visual-aware recommendation, in this paper, we propose a contextual bandit algorithm, where the convolutional neural network (CNN) is utilized to learn the reward function along with an upper confidence bound (UCB) for exploration. We also prove a near-optimal regret bound $\tilde{\mathcal{O}}(\sqrt{T})$ when the network is over-parameterized, and establish strong connections with convolutional neural tangent kernel (CNTK). Finally, we evaluate the empirical performance of the proposed algorithm and show that it outperforms other state-of-the-art UCB-based bandit algorithms on real-world image data sets.

\end{abstract}

\section{Introduction}

Online display recommendation/advertising plays an indispensable role in online business to deliver information to customers via various channels, e.g., e-commerce and news platforms. Image ads are considered as one of the most prevalent formats to catch the attention of potential customers. To maximize the Click-Through Rate (CTR), it is crucial to choose the most appealing image among candidates and display it on spot. For example, on Amazon, only one image of the product is allowed to display on the first-level page although multiple images are provided. Such a scenario can be easily found on other advertising platforms. On the other hand, the exploration-exploitation dilemma also exists in visual advertising,
as all the candidates should be displayed to customers for exploring new knowledge.


The contextual Multi-Armed Bandit (MAB) can naturally formulate the procedure and has shown success in online advertising \citep{2010contextual,wang2021hybrid, 2016collaborative,ban2021local}.  In the standard bandit setting, suppose there are $n$ arms (images) in a round, each of which is represented by a feature vector or matrix, the learner needs to pull an arm and then observe the reward (CTR). 
The goal of this problem is to maximize the expected rewards of played rounds.

The contextual MAB has been studied for decades \citep{2011improved, 2013finite, ban2021multi}. By imposing linear assumptions on the reward function, a line of works \citep{2011improved, 2014onlinecluster, ban2021local} have achieved promising results in both theory and practice, where the expected reward is assumed to be linear with respect to the arm vector.
However, this assumption may fail in real-world applications \citep{valko2013finite}.
To learn the non-linear reward function, recent works have embedded deep neural networks in the contextual bandits. \citet{zhang2020neural} and \citet{zhou2020neural} both used a fully-connected neural network to learn the reward function; the former adopted the Thompson sampling and the latter adopted the Upper Confidence Bound (UCB) strategies for exploration.


Inspired by visual recommendation, in this paper, 
we introduce a bandit algorithm, \sysn, which builds on Convolutional Neural Networks (CNN) bonding with a UCB-based exploration strategy. To best of our knowledge, we provide the first near-optimal regret bound for (Convolutional) neural bandits with Lipschitz and Smooth activation functions.
Inspired by recent advances on the convergence theory in over-parameterized neural networks\citep{du2019gradient, allen2019convergence}, we conduct analysis based on the connections among CNN, Convolutional Neural Tangent Kernel (CNTK) \citep{ntk2018neural, arora2019exact}, and ridge regression to prove an upper confidence bound and the regret bound.
To summarize, our key contributions are as follows:
\begin{enumerate}
\item We propose a new algorithm, \sysn, modeled as a convolutional neural network along with a new upper confidence bound. It focuses on capturing the visual patterns of image input.

\item Under the standard assumptions of over-parameterized networks, we provide the first regret bound for (Convolutional) neural bandits with Lipschitz and Smooth activation functions. Furthermore, we prove the equivalence of CNTKs, i.e., the dynamic CNTKs tend to be equivalent to the initialization of CNTK during the training process.

\item We conduct extensive experiments on real-world image data sets and show that \sysn achieves significant improvements on the visual advertising problem over seven baselines.
\end{enumerate}



The rest of the paper is organized as follows. After briefly introducing the related work in Section 2, we formally present the problem definition and the proposed algorithm in Sections 3 and 4, respectively. The main theorems and the proof workflow are introduced in Sections 5 and 6. In the end, we show the empirical results in Section 7 before concluding the paper in Section 8. The proofs are attached in Appendix.


\section{Related Work}

\textbf{Contextual bandits}.
The most studied literature is the linear contextual bandit \citep{2010contextual, 2011improved,2013finite}, where the reward function is governed by the product of the arm feature vector and an unknown parameter. By the UCB exploration, many algorithms \citep{2011improved, 2016collaborative} can achieve the near-optimal $\tilde{\mathcal{O}}(\sqrt{T})$ regret bound.
To break the linear assumption, \citet{filippi2010parametric} dealt with a composition of linear and non-linear function; \citet{bubeck2011x} assumed it to have a Lipschitz property in a metric space; \citet{valko2013finite} embedded the reward function into Reproducing Kernel Hilbert Space. Above assumptions all can be thought of as special cases in the problem we study. To improve bandit's performance on online recommendation, other variants of bandit have been studied such as clustering of bandits \citep{2014onlinecluster, 2016collaborative, ban2021local}, Multi-facet Bandits \citep{ban2021multi}, and outlier arm detection \citep{zhuang2017identifying, ban2020generic}.

\textbf{Neural Bandits}. Thanks to the representation power, deep neural networks (DNN) have been adapted to bandits.
\citet{riquelme2018deep} used L-layer DNN to learn a representation for each arm and applies Thompson sampling on the low-dimension embeddings for exploration.
\citet{wang2021hybrid} extended the above framework to visual advertising by using CNN to train arm embeddings. However, they did not provide the regret analysis.
\citet{zhang2020neural} first introduced a provable neural bandit algorithm in Thompson sampling where a fully-connected network learns the reward function. The most relevant work \citep{zhou2020neural} used the fully-connected neural network to learn the reward function with UCB exploration and provided a regret bound. However, \sysn is different from \citep{zhou2020neural, zhang2020neural} from many perspectives: First, \sysn targets on image data as CNN usually outperforms fully-connected network on visual recommendation according to recent advances in computer vision \citep{simonyan2014very}; Second, \citet{zhou2020neural, zhang2020neural} built the regret analysis on the recent advances in fully-connected network \citep{allen2019convergence, ntk2019generalization}, which can not directly apply to CNN. Instead, we established strong connections among CNN, CNTK, and Ridge regression.
Third, we have different assumptions on activation functions, where \sysn uses the Lipschitz-smooth function contrary to ReLU function used in \cite{zhou2020neural, zhang2020neural}; Fourth, \sysn achieved a better regret bound than them.

\section{Problem Definition}

\textbf{Notations}. 
Given the number of rounds $T$, we denote by $[T]$ the sequence $\{1, \dots, T \}$ and $\{\bx_t\}_{t=1}^{T}$ the sequence $( \bx_1, \dots, \bx_T )$. We use $\|v\|_2$ to denote the Euclidean norm for a vector $v$ and $\|\bw\|_2$ and $\| \bw  \|_F $ denote the spectral and Frobenius norm for a matrix $\bw$. We use $\langle \cdot, \cdot \rangle$ to denote the standard Euclidean inner product between two vectors or two matrices.
We use $C$ with subscripts to denote the constants and $\Psi$ with subscripts to denote the intermediate results.
We use standard notations $\oo{\cdot}$ and $\Omega( \cdot )$ to hide constants and $\mathcal{N}(\mu, 1)$ to represents the Gaussian distribution of mean $\mu$ and variance $1 $.

In standard stochastic MAB, a learner is faced with $n$ arms in each round $t \in [T]$, where each arm is represented by a vector or matrix $\bx_{t,i}, i \in [n]$. The learner aims to select an arm to maximize the reward of each round. The reward is assumed to be governed by a function with respect to $\bx_{t, i}$
\[
r_{t, i} = f^\ast(\bx_{t,i}) + \xi_{t, i},
\]
where $f^{\ast}$ is a linear or non-linear reward function satisfying  $ 0 \leq f^\ast(\bx_{t,i}) \leq 1$ and $\xi_{t,i}$ is the noise drawn from $\mathcal{N}(0, \cdot )$. For brevity, we use $\bx_t$ to denote the selected arm and $r_t$ to represent the received reward in round $t$. Following the standard evaluation of bandits, the regret of $T$ rounds is defined as 
\[
R_T = \bbe{ \sum_{t=1}^T (r_t^\ast - r_t)} = \sum_{t=1}^T \left({f^\ast(\bx^{\ast}_t) - f^\ast(\bx_t)} \right)
\]
where $\bx_t^\ast = \arg \max_{i \in [n]} f^\ast(\bx_{t, i})$. Our goal is to design a pulling algorithm to minimize $R_T$. 

\section{Proposed Algorithm}\label{sec:method}

Motivated by the applications of visual recommendation, we consider each arm as an image represented by a matrix $\bx_{t, i} \in \bbr^{c \times p}$, where $c$ is the number of input channels and $p$ is the number of pixels.   
We denote by $f_{\text{cnn}}(\bx; \btheta)$ a CNN model to learn the reward function $f^\ast$, consisting of $L$ convolutional layers and one subsequent fully-connected layer.
For any convolutional layer $l \in [L]$,  layer $l$ has the same number of channels $m$. 
For simplicity,  we use the standard zero paddings and set stride size as $1$ to ensure the output of each layer has the same size, following the setting of \citep{arora2019exact, du2019gradient}. 
Let $h^l $ be the output of layer $l$ and thus $h^l \in \bbr^{m \times p}$. For convenience, we may use $h^0$ to denote the input $\bx_{t, i}$.

To represent the convolutional operation of layer $l$, we use an operator $\phi(\cdot)$ to divide the input $h^{l-1}$ into $p$ patches, where each patch has the size $qm$. $m$ is the number of channels of last layer and $q$ is considered as the size of kernel (assume all the kernels have the same size for simplicity of analysis). For example, give a $2 \times 2$ kernel, then $q$ should set as $4$. $\phi(h^{l-1}) \in \bbr^{qm \times p} $ is figured as following:
\[
\begin{Bmatrix}
{h^{l-1}_{0, 0:3}}^\intercal, &\dots,  &{h^{l-1}_{0, p-1: p+2}}^\intercal \\
 \dots,          &\dots,      &\dots \\
{h^{l-1}_{m-1, 0:3}}^\intercal, &\dots,  &{h^{l-1}_{m-1, p-1: p+2}}^\intercal \\

\end{Bmatrix}.
\]
In accordance, we have the kernel weight matrix $\bw^l \in \bbr^{m \times qm}$. Then, the convolutional operation of layer $l$ can be naturally represented by 
$\bw^{l} \phi(h^{l-1})$.
Therefore, the output of layer $l \in [L]$ is defined as
\[
\begin{aligned}
&h^1 = \frac{1}{\sqrt{qm}} \sigma \left(\bw^1  \phi(\bx_{t, i}) \right), \  \bw^1 \in \bbr^{m \times qc} \\
& h^l = \frac{1}{\sqrt{qm}} \sigma \left(  \bw \phi_l(h^{l-1})   \right ), \  \bw^l \in \bbr^{m \times qm},   \ \text{for} \ 2 \leq l \leq L,
\end{aligned}
\]
where $\sigma$ is the activation function.
Finally, with a fully-connected layer $L+1$, the output is defined as
\[
f_{\text{cnn}}(\bx_{t,i}; \btheta) = \langle \bw^{L+1}, h^{L} \rangle   /\sqrt{m} ,
\]
where the vector $\btheta = \left( \text{vec}(\bw^1)^\intercal, \dots,  \text{vec}( \bw^{L+1})^\intercal \right)^\intercal \in \bbr^d$ represents the learned parameters. We denote by $g_{\text{cnn}}(\bx_{t,i}; \btheta) = \triangledown_{\btheta} f_{\text{cnn}}(\bx_{t,i}; \btheta) \in  \bbr^d $ the gradient of the neural network $f$.

\begin{algorithm}[t]
\renewcommand{\algorithmicrequire}{\textbf{Input:}}
\renewcommand{\algorithmicensure}{\textbf{Output:}}
\caption{ \sysn }\label{alg:main}
\begin{algorithmic}[1] 
\Require $f$,  $T, \eta, k, \lambda$ 
\State  Initialize $\btheta_0:  (\bw^{1}, \dots, \bw^{L+1}) \sim \mathcal{N}(0, 1)$ and $ \bw^{L+1} \sim  \mathcal{N}(0, 1/m) $
\State $\mathbf{A}_0 = \lambda \mathbf{I}$
\For{each $t \in [T]$}
\State Observe $n$ arms $\{\bx_{t,1}, \dots, \bx_{t, n}\}$
\For{ each $i \in [n]$}
\State 
\[
\begin{aligned}
U_{t,i} =& f_{\text{cnn}}(\bx_{t, i}; \btheta_{t-1}) + \Psi_1 \| g_{\text{cnn}}(\bx_{t, i}; \btheta_{t-1})   /\sqrt{m}  \|_{\mathbf{A}_{t-1}^{-1}}  \\
& + \Psi_2  +   \Psi_3 \text{ (defined in Theorem \ref{theo:ucb})}
\end{aligned}
\]
\EndFor
\State $\bx_t  = \arg \max_{i \in [n]}   U_{t,i}  $     \ \ \      
\State Play $\bx_{t}$ and observe reward $r_t$
\State $\mathbf{A}_t = \mathbf{A}_{t-1} +    g_{\text{cnn}}(\bx_{t}; \btheta_t)  g_{\text{cnn}}(\bx_{t}; \btheta_t)^\intercal  /m$
\State $\btheta_t$ = \text{GradientDescent}($\btheta_0$,  $\{\bx_i\}_{i=1}^t, \{ r_i\}_{i=1}^t $)
\EndFor

\\

\Procedure{GradientDescent}{$\btheta_{0}$}
\State $\btheta^{(0)} = \btheta_0$
\For{ $i \in \{1, \dots, k\}$ }
\State $\btheta^{(i)} = \btheta^{(i-1)} - \eta \triangledown_{\btheta^{(i-1)}}\mathcal{L} \left( \{\bx_i\}_{i=1}^t, \{ r_i\}_{i=1}^t \right) $
\EndFor
\State \textbf{Return} $\btheta^{(k)}$
\EndProcedure

\end{algorithmic}
\end{algorithm}

After $t$ rounds, we have $t$ selected arms $\{\bx_i\}^t_{i=1} = \{\bx_1, \dots, \bx_t\}$ and $t$ received rewards $\{r_i\}_{i=1}^t = \{r_1, \dots, r_t\}$. Thus, to learn the optimum $f^\ast$, the learning of $\btheta$ can be transform into the solution of the following minimization problem by gradient descent:
\[
\min_{\btheta} \mathcal{L} \left( \{\bx_i\}_{i=1}^t, \{ r_i\}_{i=1}^t \right)  = \frac{1}{2}\sum_{i=1}^t( f_{\text{cnn}}(\bx_i; \btheta) - r_i)^2
\]
where $\mathcal{L}$ is the quadratic loss function.

To solve the exploitation and exploration dilemma, we use the UCB-based strategy. First, we define a confidence interval for $f_{\text{cnn}}$:
\[
\bbp{|f_{\text{cnn}}(\bx_{t,i}; \btheta) - f^\ast(\bx_{t,i})| > \text{UCB}(\bx_{t, i})} < \delta,
\]
where UCB($\bx_{t, i}$) is defined in Theorem \ref{theo:ucb} and $\delta$ usually is a small constant. Then, in each round $t$, the arm is determined by 
\[
\bx_t = \arg \max_{i \in [n]} \left( f_{\text{cnn}}(\bx_{t, i}; \btheta) + \text{UCB}(\bx_{t, i}) \right)     
\]

Algorithm \ref{alg:main} depicts the workflow of \sysn. First, we initialize all parameters, where each entry of $ \bwl$ in drawn from $\mathcal{N}(0, 1)$ for $l \in [L]$ and each entry of $\bw^{L+1}$ is drawn from   $\mathcal{N}(0, 1/m)$. When observing a set of $n$ arms, \sysn chooses the arm using the UCB-based strategy. After receiving the reward, \sysn conducts the gradient descent to update $\btheta$ with the new collected training pairs $\{\bx_i\}_{i=1}^t$ and $\{r_i\}_{i=1}^t$.

\section{Main Theorems}

In this section, we introduce two main theorems, the upper confidence bound of CNN function and the regret analysis of \sysn,  inspired by recent advances in convergence theory of ultra-wide networks \citep{du2019gradient, allen2019convergence, arora2019exact} and analysis in the linear contextual bandit \citep{2011improved}.

Our analysis is based on the Lipschitz and Smooth activation function which holds for many activation functions such as Sigmoid and Soft-plus. The following condition is to guarantee the stability of training process and build connections with CNTK.

\textbf{Condition:} (Lipschitz and Smooth) Given the activation function $\sigma$, there exist a constant $\mu > 0$ such that for any $x_1, x_2 \in \bbr$, 
\begin{equation}
\begin{aligned}
&(1). |\sigma(x_1) - \sigma(x_2)| \leq \mu  |x_1 - x_2| \\
&(2). |\sigma'(x_1) - \sigma'(x_2)| \leq \mu  |x_1 - x_2|. 
\end{aligned}
\end{equation}

Before introducing main theorems, we first present the following lemma. Recall that $g_{\text{cnn}}(\bx_{t,i}; \btheta) = \triangledown_{\btheta} f_{\text{cnn}}(\bx_{t,i}; \btheta)$.
 
\begin{lemma}  \label{lemma:gounththeta}
Define
$\bg_t = [g_{\text{cnn}}(\bx_1; \btheta_t), \dots, g_{\text{cnn}}(\bx_t; \btheta_t]$ $\in \bbr^{d \times t}$ and assume $\bg_t^\intercal \bg_t \succeq  \lambda_1  \mathbf{I}$ given $\lambda_1 > 0$. 
Then, in a round $t$,  there exist $\btheta^{\ast}_t \in \bbr^{d}$ and $\bar{S} > 0$ such that for any $\bx \in \bbr^{c \times p}$ satisfying $\|\bx\|_F = 1$, it has
\[
\begin{aligned}
& (1) f^{\ast}(\bx) = \langle \ngx{t}{},  \btheta^{\ast}_t - \btheta_0 \rangle.\\
& (2) \sqrt{m}  \| \btheta^{\ast}_t - \btheta_0   \|_2 \leq \bar{S}
\end{aligned}
\]
\end{lemma}

This lemma shows than $f^{\ast}$ can be represented by a linear function with respected to the gradient $\ngx{t}{}$. Note that $\btheta^{\ast}_t$ is introduced for the sake of analysis rather than the ground-truth parameters. The we provide the following UCB.


\begin{theorem} \label{theo:ucb}
In the round $t+1$,  given a set of arms $\{ \bx_i\}_{i=1}^{t}$ and a set of rewards $\{r_i\}_{i=1}^{t}$, let $f_{\text{cnn}}(\bx; \btheta)$ be the convolutional neural network defined in Section \ref{sec:method}. Then, there exist constants $C_0 > 0$, $ 1 < C_1, C_2 <2$ such that if 
\begin{equation} \label{eq:cond}
\begin{aligned}
m &\geq  \max \left \{    \frac{ \Omega\left( t^4 (C_1\mu)^L e^{C_1 L \sqrt{q} + C_2} \right) }     {\lambda C_0},    \Omega \left[ \log \left( \frac{Lt}{\delta} \right) \right] \right \} \\
\eta & \leq ( m \lambda + 1)^{-1},   \   \ \ 
\sqrt{m} \|\btheta^{\ast}_t - \btheta_0 \|_2 \leq \bar{S}
\end{aligned}
\end{equation}
with probability at least $1-\delta$, for any $\bx \in \bbr^{c \times p}$ satisfying $\|\bx\|_F =1$, we have the following upper confidence bound:
\[
\left| f^\ast (\bx)  - f_{\text{cnn}}(\bx; \btheta_t) \right| \leq  \Psi_1 \| \ngx{t}{}  /\sqrt{m}  \|_{\mathbf{A}_t^{-1}}  + \Psi_2  +   \Psi_3.
\]
where 
\[
\begin{aligned}
&\Psi_1 = \left ( \sqrt{  \log \left(  \frac{\deter(\mathbf{A}_t)} { \deter(\lambda\mathbf{I}) }   \right)   - 2 \log  \delta }  +  \sqrt{\lambda} \bar{S}   \right) \\
& \Psi_2 = \\
& \left(  \sqrt{L+1} \left( \sqrt{p}(C_1\mu\sqrt{q})^L/m +   \sqrt{q} \Psi_{L, (k')}  ( (C_1\mu)^L + 2)   \right) \right)\\
& \cdot \bigg[     \frac{    \bar{A}_1 \cdot \sqrt{t}\Psi_3  +   m \bar{A}_2 }{ m \lambda } + \sqrt{  \frac{t}{m\lambda} }  \bigg ]    \\
& \Psi_3 =   \Big \{ C_2 (\Psi_{L, (k')} + (C_1\mu)^L ) + \\
&\sqrt{q} (1 + \Psi_{L, (k')}  ) w  \left[(L-1) ( \Psi_{L, (k')} +  (C_1\mu)^L) + 1 \right] \Big \}  /\sqrt{m}
\end{aligned}
\]
and
\[
\begin{aligned}
\bar{A}_1 = & \sqrt{t(L+1)} \sqrt{p}(C_1\mu\sqrt{q})^L/m  \\
 & + \sqrt{tq(L+1})  \Psi_{L, (k')}  ( (C_1\mu)^L + 2) \\
\bar{A}_2 = &  \lambda \sqrt{L+1}w/ \sqrt{m} 
\   \   \ \   \Psi_{L, (k')} = \frac{\mu w \left( (2 \mu  C_1 \sqrt{q})^L -1 \right) }{m ( 2\mu  C_1 \sqrt{q} -1)} \\
 w = & \frac{  2 t\sqrt{2}  \mu^{L} e^{C_1(L-1)\sqrt{q} + C_2} ( (C_1 \mu)^L + 1)  } {C_0}
\end{aligned}
\]

\end{theorem}

With above UCB, we provide the following regret bound of \sysn.

\begin{theorem}  \label{theo:regret}
After $T$ rounds, 
given the set of arms $\{ \bx_t\}_{t=1}^{T}$ and the set of rewards $\{r_t\}_{t=1}^{T}$, let $f_{\text{cnn}}(\bx; \btheta)$ be the convolutional neural network defined in Section \ref{sec:method}. Then,
there exist constants $C_0>0$, $1 < C_1, C_2 < 2$ such that if 
\[
\begin{aligned}
m &\geq  \max \left \{  \frac{ \Omega\left( T^4 (C_1\mu)^L e^{C_1 L \sqrt{q} + C_2} \right) }     {\lambda C_0},    \Omega \left[ \log \left( \frac{LT}{\delta} \right) \right] \right \} \\
\eta & \leq ( m \lambda + 1)^{-1},   \   \ \ 
\sqrt{m} \|\btheta^{\ast}_t - \btheta_0 \|_2 \leq \bar{S}, \forall t \in [T],
\end{aligned}
\]
 then with probability at least $1 -\delta$,  the regret of \sysn is upper bounded by 
\begin{equation}
\begin{aligned}
&R_T \leq    2 \sqrt{2T \bar{d} \log ( 1 + T/\lambda) + 1}   \cdot\\
&\left( \sqrt{ \bar{d} \log ( 1 + T/\lambda) + 2\log(1/\delta)+1} +  \sqrt{\lambda} \bar{S} \right) + 2.
\end{aligned}
\end{equation}
where
\begin{equation} 
\begin{aligned}
&\bar{d}  =  \frac{\log \det \left( \mathbf{I} +  \bg^\intercal_0  \bg_0 /\lambda  \right)  }{ \log (1 + T\lambda)}  \\
&\bg = ( \ngx{0}{1}, \dots, \ngx{0}{T}) \in \bbr^{d \times T}.
\end{aligned}
\end{equation}

\end{theorem}
Note that as $\bar{S}$ is fixed to every arm in round $t$ and it can be calculated according to Lemma \ref{lemma:gounththeta}, it does not affect the exploration performance of UCB and thus the complexity of regret bound.
$\bar{d}$ is the effective dimension first defined in \citet{valko2013finite} to analyze the kernel bandits and then applied by \citet{zhou2020neural} to measure the diminishing rate of NTK \citep{ntk2018neural}. We adapt it to the CNTK \citep{arora2019exact, yang2019scaling} to alleviate the predicament caused by the blowing up of parameters.  

\para{Comparison with existing works.} The above regret bound can be reduced to the time complexity of $\oo{ \sqrt{T  } \oo{\log \oo{ T}}}$, which is the same as the state-of-the-art regret bound in linear contextual bandits \citep{2011improved} while we do not impose any assumption on the reward function. The most relevant works  \citep{zhou2020neural, zhang2020neural}, instead, achieve the regret bound of 
\begin{equation} 
\begin{aligned}
\mathcal{O} \Bigg( \sqrt{T \oo{\log \oo{ T}}} \cdot   \Big( \sqrt{\oo{\log \oo{ T}}} \\ 
+ \oo{T} \cdot (1- \oo{T^{-1}})^k \Big ) \Bigg).
\end{aligned}
\end{equation}
 The term $ \oo{T} \cdot (1- \oo{T^{-1}})^k $ can result in the exploding of error. To eliminate this error, they need an additional assumption on $k$ that needs to be extremely large, while \sysn does not have this concern.
 
 Moreover, to achieve the regret bound, we require $m \geq \Omega(T^4 e^{L})$ while they require $m \geq \Omega( T^{24} L^{21})$. In practice, $T$ usually is much larger than $L$. For example, commonly used CNN models such as VGG (L=16) \citep{simonyan2014very} are not extremely deep. Instead, $T \geq 1 \times 10^5$ in the experiments of many bandit papers \citep{zhou2020neural, 2010contextual, li2016contextual}.


\section{CNTK and Main Proof}

In this section, we present proof's sketch  of Theorem \ref{theo:regret} with important lemmas.

\para{Equivalence to Convolutional Neural Tangent Kernel}.
First, we introduce the lemma to bridge CNN and CNTK. 
Neural Tangent Kernel (NTK) \citep{ntk2018neural, allen2019convergence, arora2019exact} usually is defined as the feature space formed by the gradient at random initialization. Therefore, given any two arms $\bx_1, \bx_2$, CNTK is defined as
\[
K^{\text{CNTK}}(\bx_1, \bx_2) = \langle g_{\text{cnn}}(\bx_1; \btheta_0),   g_{\text{cnn}}(\bx_2; \btheta_0)   \rangle .
\]
Given an arm $\bx$, the CNTK objective is defined as
\[
f^{\text{CNTK}}(\bx; \btheta_t -\btheta_0) = \langle   g_{\text{cnn}}(\bx; \btheta_0), \btheta_t - \btheta_0 \rangle.
\]
\citet{ntk2018neural} prove that, for a fully-connected network, the dynamic NTKs are identical to NTK at initialization  when $m$ is infinite, because $\lim_{m \rightarrow \infty} \| \btheta_t - \btheta  \|_2 = 0 $. We present the following lemma to show that the same results hold for CNTK.

\begin{lemma} [\textbf{CNTK}]
In a round $t$, with probability at least
$1 - \delta$, if $m$ satisfies Condition  Eq. (\ref{eq:cond}),  we have
\[
\begin{aligned} \label{lemma:CNTK}
(1)   & \   \| g_{\text{cnn}}(\bx; \btheta_t)- g_{\text{cnn}}(\bx; \btheta_0)\|_2  \\
 \leq   & \sqrt{q(L+1}) ( (C_1\mu)^L + 2)   \frac{\mu w \left( (2 \mu  C_1 \sqrt{q})^L -1 \right) }{m ( 2\mu  C_1 \sqrt{q} -1)}   \\
 (2) & \  | \langle g_{\text{cnn}}(\bx_1; \btheta_t),   g_{\text{cnn}}(\bx_2; \btheta_t) \rangle - K^{\text{CNTK}}(\bx_1, \bx_2) |  \\ 
 \leq & (L+1) \left( 2\sqrt{p}(C_1\mu\sqrt{q})^L/m +   \sqrt{q} \Psi_{L, (k')}  ( (C_1\mu)^L + 2)   \right)  \\
 & \cdot \sqrt{q} \Psi_{L, (k')}  ( (C_1\mu)^L + 2)  \\
 (3) & \  |f_{\text{cnn}}(\bx; \btheta_t)  - f^{\text{CNTK}}(\bx; \btheta_t  -\btheta_0)|   \\
\leq & \Big( \Psi_{L, (k')} (C_2  + 1)  +  (C_1\mu)^L C_2 \\
 & + \sqrt{q} w((L-1)  (C_1\mu)^L  + 1)  \Big) /\sqrt{m}.
\end{aligned}
\]
\end{lemma}
The above lemma shows that $f_{\text{cnn}}(\bx; \btheta_t) \rightarrow  f^{\text{CNTK}}(\bx; \btheta_t -\btheta_0) $ when $ m \rightarrow \infty$.

Next, we present the following lemma to connect CNTK with ridge regression. 
According to the linear bandit \citep{2011improved}, the estimation for $\btheta$ by standard ridge regression is defined as
\[
 \whtt = \mathbf{A}^{-1}_t\mathbf{b}_t  \ \  \  \  \mathbf{b}_t = \sum_{i=1}^t r_t \ngx{t}{t}/\sqrt{m}. 
\]

\begin{lemma}\label{lemma:hattbound}
In round $t$, with probability at least $1-\delta$, if $m, \eta$ satisfy Condition Eq. (\ref{eq:cond}),  we have
\[
\begin{aligned}
& \| \btheta_t - \btheta_0  -  \widehat{\btheta}_t /\sqrt{m} \|_2  \leq 
\frac{    \bar{A}_1 \cdot \sqrt{t}\Psi_3  +   m \bar{A}_2 }{ m \lambda }  + \sqrt{  \frac{t}{m\lambda} }
\end{aligned}
\]
\end{lemma} 

Based on the  above lemmas, we can bound $|f^{\text{CNTK}}(\bx; \btheta_t -\btheta_0) - \langle g_{\text{cnn}}(\bx; \btheta_0),    \widehat{\btheta}_t /\sqrt{m} \rangle   | $.
With the above lemmas, we can easily prove Theorem \ref{theo:ucb} and  \ref{theo:regret}.

\begin{lemma} \label{lemma:logabound}
With probability at least $1-\delta$, if $m, \eta$ satisfy Condition Eq. (\ref{eq:cond}),
$
\log \left( \frac{\mathbf{A}_T}{\lambda \mathbf{I}}\right)  \leq  \bar{d} \log (1 + T / \lambda) + 1.
$ 
\end{lemma}

\textbf{Proof of Theorem \ref{theo:regret}}.
For a round $t$, with probability at least $1-\delta$, its regret is
\[
\begin{aligned}
R_t &= f^\ast(\bx^\ast_t) - f^\ast (\bx_t) \\
&\leq | f^\ast (\bx^\ast_t) - f_{\text{cnn}}(\bx^\ast; \btheta_{t-1})| +   f_{\text{cnn}}(\bx^\ast_t; \btheta_{t-1})  - f^\ast(\bx_t) \\
&\leq \ucb(\bx_t^\ast) + f_{\text{cnn}}(\bx^\ast_t; \btheta_{t-1}) - f^\ast(\bx_t) \\
& \leq  \ucb(\bx_t) + f_{\text{cnn}}(\bx_t; \btheta_{t-1}) -  f^\ast(\bx_t) \\
&\leq 2 \ucb(\bx_t) 
\end{aligned}
\]
where the third inequality is as the result of pulling criteria of \sysn satisfying
\[
\ucb(\bx_t^\ast) + f_{\text{cnn}}(\bx^\ast_t; \btheta_{t-1}) \leq  \ucb(\bx_t) + f_{\text{cnn}}(\bx_t; \btheta_{t-1}).
\]
Therefore, for $T$ rounds, we have
\[
\begin{aligned}
R_T &= \sum_{t=1}^T R_t \leq 2\sum_{t=1}^T  \ucb(\bx_t) \\
 &= 2\sum_{t=1}^T  \left(   \Psi_1  \| \ngx{t-1}{t}  /\sqrt{m}  \|_{\mathbf{A}_t^{-1}}  + \Psi_2 + \Psi_3    \right) \\ 
\leq & 2  \Psi_1  \sqrt{T \sum_{t=1}^{T}  \| \ngx{t-1}{t}  /\sqrt{m}  \|_{\mathbf{A}_t^{-1}}^2     } +  2\sum_{t=1}^{T}  \Psi_2 +  2\sum_{t=1}^{T}  \Psi_3 \\ 
\leq & \underbrace{2  \Psi_1}_{\mathbf{I}_1} 
\underbrace{\sqrt{2T\log \left( \frac{\text{det}(\mathbf{A}_T)}{\text{det}(\lambda \mathbf{I})}\right) }}_{\mathbf{I}_2}  +  \underbrace{2\sum_{t=1}^{T}  \Psi_2}_{\mathbf{I}_3} +  \underbrace{2\sum_{t=1}^{T}  \Psi_3}_{\mathbf{I}_4}.
\end{aligned}
\]
where the last inequality is based on the Lemma 11 in \cite{2011improved}. For $\mathbf{I}_2$, applying Lemma \ref{lemma:logabound}, we have
\begin{equation}
\mathbf{I}_2 \leq \sqrt{2T \bar{d} \log ( 1 + T/\lambda) + 1}.
\end{equation}
For $\mathbf{I}_1$, applying lemma \ref{lemma:logabound} again, we have 
\begin{equation}
2 \mathbf{I}_1 \leq 2\left( \sqrt{ \bar{d} \log ( 1 + T/\lambda) + 2\log(1/\delta)+1} +  \sqrt{\lambda} \bar{S} \right).
\end{equation}
For $\mathbf{I}_3$ and $\mathbf{I}_4$, as the choice of $m$, we have
\begin{equation}
2\sum_{t=1}^{T}  \Psi_2  \leq 1,  \  \ \ \   2\sum_{t=1}^{T}  \Psi_3 \leq 1.
\end{equation}
Therefore, adding everything together completes the proof.

\section{Experiments}

In this section, we evaluate the empirical performance of \sysn compared with seven strong baselines on image data sets.

\para{Image data sets}. 
We choose three well-known image data sets: Mnist \citep{lecun1998gradient}, Notmnist, and Cifar-10 \citep{krizhevsky2009learning}. All of them are 10-class classification data sets. Following the evaluation setting of existing works \citep{zhou2020neural, 2013finite, deshmukh2017multi}, transform the classification into bandit problem. Consider an image $\bx \in \bbr^{c \times p}$, we aim to classify it from $10$ classes. Then, in each round, $10$ arms is presented to the learner, formed by $10$ tensors in sequence $\bx_1 = (\bx, \mathbf{0}, \dots, \mathbf{0}), \bx_2 = (\mathbf{0}, \bx, \dots, \mathbf{0}), \dots, \bx_{10} = (\mathbf{0}, \mathbf{0}, \dots, \mathbf{\bx}) \in \bbr^{10 \times c \times p} $, matching  the $10$ classes. The reward is defined as $1$ if the index of selected arm equals the index of $\bx$' ground-truth class; Otherwise, the reward is $0$. For example, an image with number "6" belonging to the $7$-th class on Mnist data set will be transformed into $10$ arms in a round and the reward will be $1$ if selecting the $7$-th arm; Otherwise, the reward is $0$.  
For Mnist and Notmnist, we transform them into a $10$-arm
bandit problem. For Cifar-10, we tranform it into a $3$-arm bandit problem to alleviate the huge computation cost caused by the input dimensions. Specifically, the arm $0$ $(\bx, \mathbf{0}, \mathbf{0})$ matches the image classes $0-3$;  the arm $1$ $( \mathbf{0},\bx, \mathbf{0})$ matches the image classes $4-7$;  the arm $1$ $(\mathbf{0}, \mathbf{0}, \bx, )$ matches the image classes $8-9$.

\para{Yelp data set \footnote{https://www.yelp.com/dataset}}. 
Yelp is a data set released in the Yelp data set challenge, which consists of 4.7 million rating entries for  $1.57 \times 10^5$ restaurants by $1.18$ million users. 
We build the rating matrix by choosing the top $2000$ users and top $10000$ restaurants and use singular-value decomposition (SVD) to extract the $10$-dimension feature vector for each user and restaurant.
In this data set, the bandit algorithm is to choose the restaurants with bad ratings. We generate the reward by using the restaurant's gained stars scored by the users. In each rating record, if the user scores the restaurant less than 2 stars (5 stars totally), the reward  is $ 1$; Otherwise, the reward is $ 0$. 
In each round,  we set $10$ arms as follows: we randomly choose one rating with reward $1$ and randomly pick the other $9$ restaurants with $0$ rewards; then, the representation of each arm is the concatenation of corresponding user feature vector and restaurant feature vector.

\textbf{Baselines}. To comprehensively evaluate \sysn's empirical performance, we choose and design seven strong baselines. 
(1) LinUCB \citep{2010contextual} assumes the reward is a product of arm feature vector and an unknown parameter and then uses ridge regression to do the estimation and UCB-based exploration;
(2) KernelUCB \citep{2013finite} uses a predefined kernel matrix to learn reward function coming with a UCB exploration strategy; 
(3) NeuralUCB \citep{zhou2020neural} uses a fully-connected neural network to learn reward function with the UCB exploration strategy; 
(4) NeuralTS \citep{zhang2020neural} applies a Thompson Sampling exploration strategy on a fully-connected neural network;
(5) NeuralEpsilon is a fully-connected neural network embedded in the $\epsilon$-greedy exploration approach.
(6) CNN\_Epsilon is the convolutional neural network with $\epsilon$-greedy exploration approach.
(7) CNN+TS is the convolutional neural network added with an intuitive Thompson Sampling method.

\textbf{Configuration details} are presented in Appendix, due to the limit of space.

\begin{figure}[t] 
    \includegraphics[width=1.0\columnwidth]{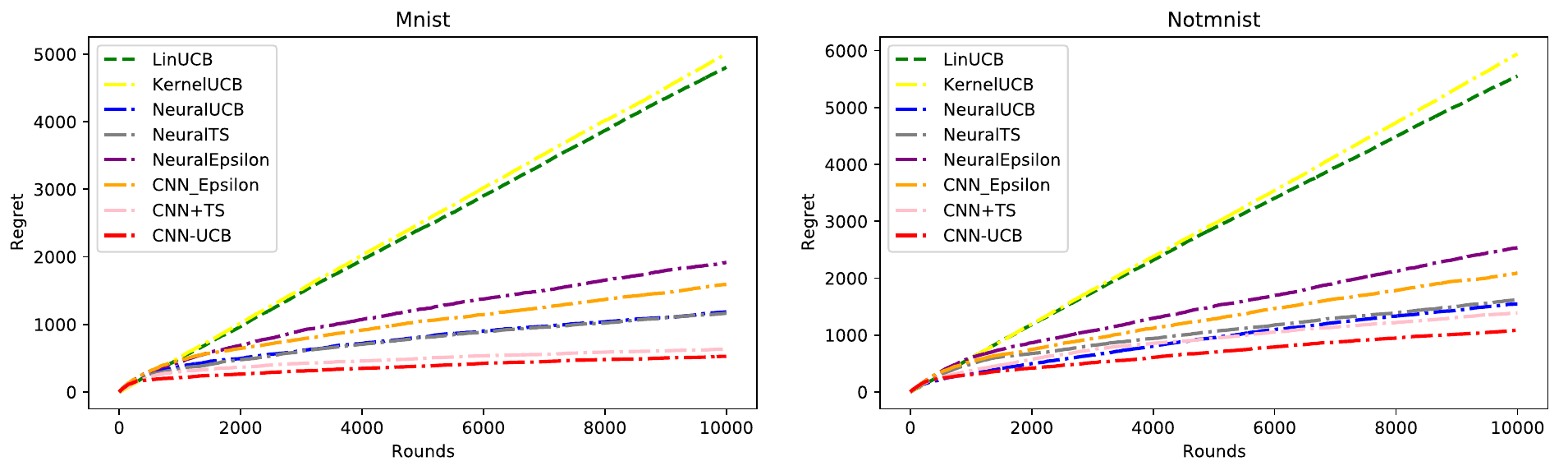}
    \centering
    \vspace{-0.5cm}
    \caption{ Regret comparison on Mnist and NotMnist. Our approach (\textbf{red line}), \sysn, outperforms all baselines. }
       \label{fig:1}
\end{figure}

\begin{figure}[t] 
    \includegraphics[width=1.0\columnwidth]{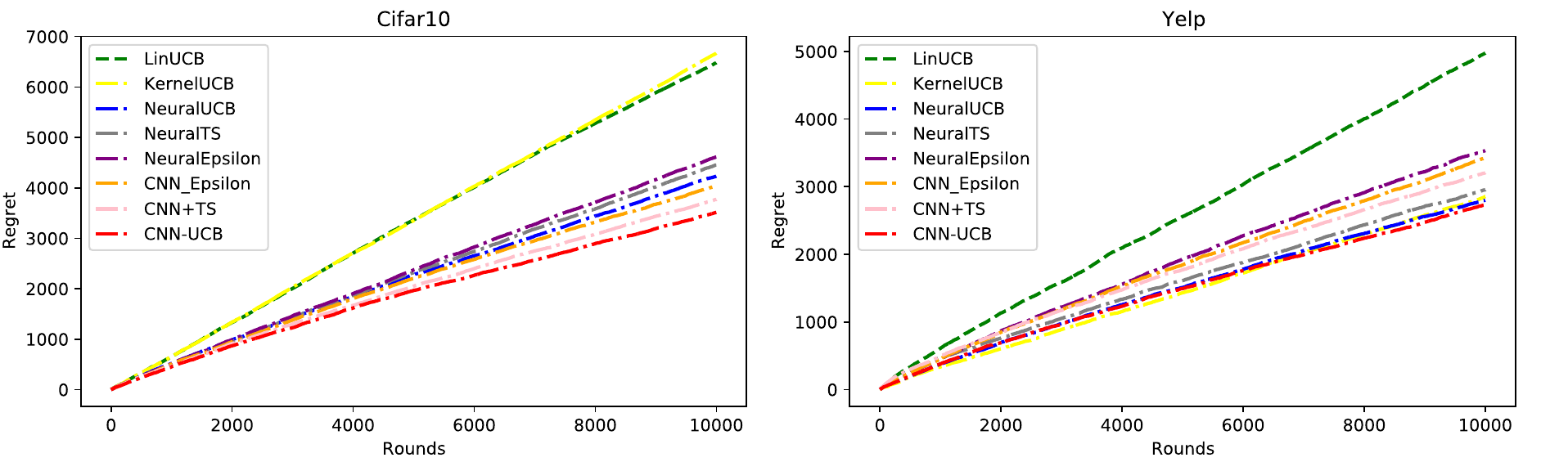}
    \centering
    \caption{ Regret comparison on .  Our approach (\textbf{red line}), \sysn, outperforms all baselines. }
       \label{fig:2}
\end{figure}

\textbf{Results}. Figure \ref{fig:1} and Figure \ref{fig:2} show the regret comparison for all algorithms on four data sets. \sysn achieves the best performance across all data sets as expected, because CNN can capture the visual pattern of image to exploit the past knowledge and our derived UCB can explore these new arms to gain new information.
For KernelUCB, it shows that a simple kernel like radial basis function has the limitation to learn complicated reward functions. LinUCB obtains the worst performance because the linear-reward assumption and a large number of input dimensions make it hard to estimate the reward function accurately.
Neural methods obtain the lower regret, such as NeuralUCB and NeuralTS, outperforming LinUCB and KerUCB, thanks to the representation power of neural network.
CNN+TS obtains the best performance on three Image data sets among baselines because of the superiority of CNN. However, our derived UCB usually makes better exploration decisions compared to an intuitive TS and the analysis of CNN+TS still is a vacancy.

\section{Conclusion}

In this paper, we propose a contextual bandit algorithm, \sysn, which uses the CNN to learn the reward function with an UCB-based exploration.
We also achieve an near-optimal regret bound $\tilde{\mathcal{O}}(\sqrt{T})$  built on the connections among CNN, CNTK, and ridge regression. 
\sysn has direct applications on visual-aware recommendation and outperforms state-of-the-art baselines on real-world image data sets.

\clearpage

\bibliography{ref}

\newpage 
\appendix

\onecolumn

\para{Configurations}. For LinUCB, following \citep{2010contextual}, there is a exploration constant $\alpha$ (to tune the scale of UCB) and we do a grid search for $\alpha$ over $(\mathbf{0.01}, 0.1, 1)$. For KernelUCB \citep{2013finite}, we use the radial basis function kernel and stop adding contexts after 2000 rounds. There are regularization parameter $\lambda$ and exploration parameter $\nu$ in KernelUCB and we do the grid search for $\lambda$ over $(\mathbf{0.1}, 1, 10)$ and for $\nu$ over $(0.01, \mathbf{0.1}, 1)$. For NeuralUCB and NeuralTS, following setting of \citep{zhou2020neural, zhang2020neural}, we use a 2 fully-connected layer with the width $100$ and conduct the grid search for the exploration parameter $\nu$ over $(\mathbf{0.001}, 0.01, 0.1, \mathbf{1})$ and for the regularization parameter $\lambda$ over $(0.01, \mathbf{0.1}, 1)$. For NeuralEpsilon, we use the same neural network with NeuralUCB/TS and do the grid search for the exploration probability $\epsilon $ over $(0.01, \mathbf{0.1}, 0.2)$. For CNN-UCB, we use two convolutional layers connected with two fully-connected layers, where the first convolutional layer has $32$ channels and the second have $64$ channels. For image data sets, we use the 2-dimension CNN while using 1-dimension CNN for Yelp data set. And we conduct the grid search for the exploration parameter $\nu$ over $(0.001, \mathbf{0.01}, 0.1, 1)$ and for the regularization parameter $\lambda$ over $(1 \times 10^{-3} , 1 \times 10^{-4}, \mathbf{1 \times 10^{-5}})$.
For CNN+TS, we adapt Thompson Sampling to the the same neural network structure with CNN-UCB where the variance is $\nu \| \ngxt \|_2^2$. And we conduct the grid search for its exploration parameter $\nu$ over $(0.001, \mathbf{0.01}, 0.1, 1)$ and for its regularization parameter $\lambda$ over $(1 \times 10^{-3} , \mathbf{1 \times 10^{-4}}, 1 \times 10^{-5})$.
For CNN-Epsilon, we use the same CNN with CNN-UCB and do the grid search for the exploration probability $\epsilon $ over $(0.01, \mathbf{0.1}, 0.2)$. For the neural bandits including NeuralUCB/TS and CNN-UCB, as it has expensive computation cost to store and compute the whole matrix $\mathbf{A}_t$, we use a diagonal matrix which consists of the diagonal elements of $\mathbf{A}_t$ to approximate $\mathbf{A}_t$. For all grid-searched parameters, we choose the best of them for the comparison and report the averaged results of $5$ runs.

\newcommand{\gxt}{g(\mathbf{x}_t; \btheta_t)}
\newcommand{\gx}[3][2]{g(\mathbf{x}_{#3}; \btheta_{#2})}

\section{Main Proofs}

For the simiplification of notations, we use $f(\cdot; \cdot), g(\cdot; \cdot)$ to represent $f_{\text{cnn}}(\cdot; \cdot), g_{\text{cnn}}(\cdot; \cdot)$.

\noindent \textbf{Proof of Theorem \ref{theo:ucb} }.
Given an arm $\bx_t$,  based on the Lemma \ref{lemma:gounththeta},  we have
\[
\begin{aligned}
 &\left| \hx  - \fx      \right| \\
\leq &    \left|  \left \langle  \gx{t}{t} /\sqrt{m},  \sqrt{m} (\bts - \btheta_0) \right \rangle   -   \left \langle  g(\bx_i, \btheta_t)/\sqrt{m},  \whtt \right \rangle \right | + \left| \fx -  \langle \gx{t}{t}/\sqrt{m} , \whtt \rangle \right|
\end{aligned}
\]
Then, based on the Theorem 2 in \citet{2011improved}, we have
\[
\begin{aligned}
 &   \left|  \left \langle  \gx{t}{t} /\sqrt{m},  \sqrt{m} (\bts  -  \btheta_0 ) \right \rangle   -   \left \langle  g(\bx_i, \btheta_t)/\sqrt{m},  \whtt \right \rangle \right |
  \\
& \leq \left ( \sqrt{  \log \left(  \frac{\deter(\mathbf{A}_t)} { \deter(\lambda\mathbf{I}) }   \right)   - 2 \log  \delta }  + \lambda^{1/2} \bar{S}   \right) \| \gx{t}{t}  /\sqrt{m}  \|_{\mathbf{A}_t^{-1}} = \Psi_1  \| \gx{t}{t}  /\sqrt{m}  \|_{\mathbf{A}_t^{-1}}
\end{aligned}
\]
Next, we have
\[
\begin{aligned}
& \left| \fx -  \langle \gx{t}{t}/ \sqrt{m}, \whtt \rangle  \right|  \\
\leq&  \left| \fx -  \langle \gx{t}{t},   (\btheta_t - \btheta_0)        \rangle     \right| +  \left| \langle \gx{t}{t},   (\btheta_t - \btheta_0)   \rangle    -   \langle \gx{t}{t}/ \sqrt{m}, \whtt \rangle   \right|  \\
= & \underbrace{ \left| \fx -  \langle  \gx{t}{t} ,   (\btheta_t - \btheta_0)        \rangle     \right|}_{\mathbf{I}_1} + \underbrace{ | \langle \gx{t}{t},   (\btheta_t - \btheta_0 - \whtt/\sqrt{m})   \rangle   |}_{\mathbf{I_2}}  \\
\end{aligned}
\]
By Lemma \ref{lemma:f-gk}, we can prove $\mathbf{I}_1 \leq \Psi_3 $.  For $\mathbf{I_2}$, we have
\[
\begin{aligned}
&\left| \langle \gx{t}{t},   (\btheta_t - \btheta_0 - \whtt/\sqrt{m})   \rangle   \right| \\
\leq& \left(  \| \gx{0}{t}\|_2  + \|   \gx{t}{t} - \gx{0}{t}\|_2  \right)   \| (\btheta_t - \btheta_0 - \whtt/\sqrt{m})    \|_2
\leq \Psi_2.
\end{aligned}
\]
Adding everything together, the proof is completed:
\[
\left| \hx  - \fx      \right| \leq  \Psi_1  \| \gx{t}{t}  /\sqrt{m}  \|_{\mathbf{A}_t^{-1}}  + \Psi_2  + \Psi_3.
\]

\qed

\textbf{Proof of Lemma 5.1}.
Given an arm $\bx_t$ and its ground-truth reward $f^\ast(\bx_t)$, combined with previous pair $\{\bx_i\}_{i=1}^{t-1}$ and $\{r_i\}_{i=1}^{t-1}$, we define $\bg_t = [g(\bx_1; \btheta_t), \dots, g(\bx_t; \btheta_t] \in \bbr^{d \times t}$ and $\mathbf{f}^\ast = [ f^{\ast}(\bx_1),  \dots,  f^{\ast}(\bx_t) ]^\intercal \in \bbr^{t}.$

Suppose the single value decomposition of  $\bg_t$ is $ \mathbf{U} \boldsymbol{\Sigma} \mathbf{T}^{\intercal}$, where $\mathbf{U} \in \bbr^{d \times t},  \boldsymbol{\Sigma} \in \bbr^{t \times t}$, and $ \mathbf{T} \in \bbr^{t \times t}$. 
Thus, there exist $\bts = \btheta^{(0)}  + \mathbf{U} \boldsymbol{\Sigma}^{-1} \mathbf{T}^{\intercal} \mathbf{f}^\ast /\sqrt{m} $ such that
\[
\bg_t^{\intercal} \sqrt{m} (\bts - \btheta^{(0)}) = \mathbf{T} \boldsymbol{\Sigma}    \mathbf{U}^{\intercal} \mathbf{U} \boldsymbol{\Sigma}^{-1} \mathbf{T}^{\intercal} \mathbf{f}^\ast  = \mathbf{f}^\ast.
\]
This indicates for any $\bx \in \bxst$, $\langle g(\bx; \btheta_t),  \bts - \btheta_0 \rangle = f^\ast(\bx)$. Next, easy to have
\[
 m\| \bts  -\btheta_0  \|_2^2 = {\mathbf{f}^\ast}^{\intercal} \mathbf{T} \boldsymbol{\Sigma}^{-1}    \mathbf{U}^{\intercal} \mathbf{U} \boldsymbol{\Sigma}^{-1} \mathbf{T}^{\intercal} \mathbf{f}^\ast  = {\mathbf{f}^\ast}^\intercal (\bg_t^\intercal \bg_t)^{-1} \mathbf{f}^\ast 
\]
Suppose $\bg_t^\intercal \bg_t \succeq  \lambda_1  \mathbf{I}$, then we have
\[
 m \| \bts  -\btheta_0  \|_2^2 \leq  \frac{1}{\lambda_1} {\mathbf{f}^\ast}^\intercal \mathbf{f}^\ast .
\]

\qed

\textbf{Prove of Lemma 6.3}.

Define $\bg_t = [g(\bx_1; \btheta_t)/\sqrt{m}, \dots, g(\bx_t; \btheta_t/\sqrt{m}] \in \bbr^{p \times t}$. 

\[
\begin{aligned}
\log \left( \frac{\mathbf{A}_T}{\lambda \mathbf{I}}\right)
=& \log \det \left( \mathbf{I} +  \sum_{t =1}^T \gx{t}{t} \gx{t}{t}^{\intercal}/ m \lambda\right) \\
= &\log \det \left(\mathbf{I} +  \bg^\intercal_t \bg_t /\lambda \right) \\
= & \log \det \left(\mathbf{I} +\bg^\intercal_0  \bg_0 /\lambda +  (\bg^\intercal_t \bg_t - \bg^\intercal_0 \bg_0) / \lambda\right) \\
\leq &  \log \det \left( \mathbf{I} +  \bg^\intercal_0  \bg_0 /\lambda  \right) + \langle  ( \mathbf{I} +  \bg^\intercal_0  \bg_0 /\lambda )^{-1} ,    (\bg^\intercal_t \bg_t - \bg^\intercal_0 \bg_0  )/ \lambda \rangle   \\ 
\leq & \log \det \left( \mathbf{I} +  \bg^\intercal_0  \bg_0 /\lambda  \right) + \|  ( \mathbf{I} +  \bg^\intercal_0  \bg_0 /\lambda )^{-1}\|_2  \| \bg^\intercal_t \bg_t - \bg^\intercal_0 \bg_0 \|_F / \lambda \\
\leq & \log \det \left( \mathbf{I} +  \bg^\intercal_0  \bg_0 /\lambda  \right) +  \| \bg^\intercal_t \bg_t - \bg^\intercal_0 \bg_0 \|_F /\lambda\\
\leq & \log \det \left( \mathbf{I} +  \bg^\intercal_0  \bg_0 /\lambda  \right)  +  \frac{\Psi_{\mathbf{G'}}}{m} \\
\leq & \bar{d} \log (1 + T\lambda) + 1
\end{aligned}
\]
where the first inequality is due to the concavity of $\log \det$ and the last inequality is because of the definition of $\bar{d}$ and the choice of $m$. The third inequality is because $ \mathbf{I} +\bg^\intercal_0  \bg_0 /\lambda  \geq \mathbf{I}$. The fourth inequality is because of Lemma \ref{lemma:abound}.

\qed

\begin{lemma} \label{lemma:abound}
With probability at least $1-\delta$, if $m, \eta$ satisfy Condition Eq. (\ref{eq:cond}), we have
\[
\begin{aligned}
(1) &\|\mathbf{A}_t\|_2 \leq \lambda + T(L+1) \left( \sqrt{p}(C_1\mu\sqrt{q})^L/m +   \sqrt{q} \Psi_{L, (k')}  ( (C_1\mu)^L + 2)   \right)^2/m \\
(2) & \| \bg_t^{\intercal} \bg_t  - \bg_0^{\intercal} \bg_0\|_F  \leq
\frac{ \Psi_{\mathbf{G'}} }{m} \\
&= \frac{T(L+1)}{m} \left( 2\sqrt{p}(C_1\mu\sqrt{q})^L/m +   \sqrt{q} \Psi_{L, (k')}  ( (C_1\mu)^L + 2)   \right)   \sqrt{q} \Psi_{L, (k')}  ( (C_1\mu)^L + 2) 
\end{aligned}
\]

\end{lemma}

\begin{prove}
For (1) we have
\[
\begin{aligned}
\|\mathbf{A}_t\|_2 &=  \|\lambda \mathbf{I} + \sum_{i=1}^{t} \gx{t}{i} \gx{t}{i}^\intercal /m\|_2\\
& \leq  \lambda  + \sum_{i=1}^{t}\|  \gx{t}{i}   \|_2^2/m \\
&\leq \lambda + t(L+1) \left( \sqrt{p}(C_1\mu\sqrt{q})^L/m +   \sqrt{q} \Psi_{L, (k')}  ( (C_1\mu)^L + 2)   \right)^2/m
\end{aligned}
\]
where the last inequality is because of Lemma \ref{lemma:CNTKs}.
For (2)  we have
\[
\begin{aligned}
&\| \bg_t^{\intercal} \bg_t  - \bg_0^{\intercal} \bg_0\|_F = \frac{1}{m} \sqrt{ \sum_{i}^T  \sum_{j}^T  |\langle \gx{t}{i} , \gx{t}{j}    \rangle -\langle  \gx{0}{i} , \gx{0}{j}   \rangle |^2  }\\
\leq &  \frac{1}{m} \sqrt{ \sum_{i}^T  \sum_{j}^T   \left(  \left(  \|\gx{t}{i}\|_2 + \| \gx{0}{j}\|_2      \right)  \|\gx{t}{j} - \gx{0}{i}\|_2   \right)^2 }  \\
\leq & \frac{1}{m} \sqrt{ \sum_{i}^T  \sum_{j}^T \left(  \sqrt{L+1} \left( 2\sqrt{p}(C_1\mu\sqrt{q})^L/m +   \sqrt{q} \Psi_{L, (k')}  ( (C_1\mu)^L + 2)   \right)  \sqrt{q(L+1})  \Psi_{L, (k')}  ( (C_1\mu)^L + 2)   \right)^2} \\
= & \frac{T(L+1)}{m} \left( 2\sqrt{p}(C_1\mu\sqrt{q})^L/m +   \sqrt{q} \Psi_{L, (k')}  ( (C_1\mu)^L + 2)   \right)   \sqrt{q} \Psi_{L, (k')}  ( (C_1\mu)^L + 2)    
\end{aligned}
\]
where second inequality is based on Lemma \ref{lemma:CNTKs}.
The proof is completed.

\end{prove} 

\section{Proof of Lemma 6.1}

By the definition of convolutional operation $\pphi{\cdot}$, we have
\[
\begin{aligned}
&\|h^{l-1}\|_F \leq \| \phi_l(h^{l-1}) \|_F \leq \sqrt{q} \| h^{l-1}  \|_F \\
&\|\pphi{h^{l-1}}  - \pphi{{h^{l-1}}'}\|_F \leq \sqrt{q} \|h^{l-1} - {h^{l-1}}'\|_F.
\end{aligned}
\]

For any matrix $h^{l-1} \in \bbr^{m \times p}$, for the sake of presentation,  the operator $\phi(\cdot)$ is represented by
\[
\phi(h^{l-1}) = h^{l-1} \diamond \bbk \in \bbr^{mq\times p}.
\]
For each $l \in {1, \dots, L}$, the activation function $\sigma(\cdot)$  is represented by
\[
\sigma( \bwl \pphi{h^{l-1}}) = \bbd^l \odot \left( \bwl \pphi{h^{l-1}}  \right),
\]
where $\bbd^l \in \bbr^{m \times p}$ and
\[
\begin{Bmatrix}
\sigma'\left( \left[ h^l \pphi{h^l}  \right]_{0,0} \right), &\dots,  & \sigma' \left(\left[ h^l \pphi{h^l}  \right]_{0,p-1} \right) \\
 \dots,          &\dots,      &\dots \\
\sigma' \left( \left[ h^l \pphi{h^l}  \right]_{m-1,0} \right) , &\dots,  & \sigma' \left( \left[ h^l \pphi{h^l}  \right]_{m-1, p-1}\right) \\
\end{Bmatrix}.
\]

Therefore,
given the set $\{\bx_i\}_{i=1}^t$ and $\{ r_i \}_{i=1}^t$,
 the formula of the gradient of CNN $f$ with respect to one layer $\bw^l$ is derived as :
\[
\frac{ \partial L (\btheta)}{\partial \bw^l} = \sum_{i=1}^t(f(\bx_i; \btheta) - r_i ) \phi(h^{l-1}) {\bw^{L+1}}^\intercal  \left( \prod_{j=l+1}^L  \odot \bbd^{j}  \bw^j (h^{j-1} \diamond \bbk) \right) \odot \bbd^l.
\]
We omit the subscript $i$ for the brevity.

\subsection{Lemmas}
\textbf{Lemma 9.0} [Theorem 7.1 in \cite{du2019gradient}]
\textit{
In a round $t$, at $k$-th iteration,  let $\mathbf{F}^{(k)}_t = \left( f(\bx_1; \btheta^{(k)}), \dots, f(\bx_t; \btheta^{(k)})\right)^{\intercal}$ and $ \mathbf{R}_t = \left( r_1, \dots, r_t  \right)^\intercal$. There exists a constant $C_0$, such that for any $k' \in [k]$, it has
\[
\|\mathbf{R}_t - \mathbf{F}_t^{(k')}\|^2_2  \leq (1 - C_0 \eta)^{k'} \ \| \mathbf{R}_t - \mathbf{F}_t^{(0)}\|^2_2,
\]
where $C_0 \eta < 1$.
}

Therefore, we have 
\[ \|\mathbf{R}_t - \mathbf{F}_t^{(k)}\|_2 \leq  (1 - \lambda \eta)^{k/2} \| \mathbf{R}_t - \mathbf{F}_t^{(0)}\|_2 \leq   \| \mathbf{R}_t - \mathbf{F}_t^{(0)}\|_2 \leq \sqrt{2t},
\]
where the last inequality is because $r_t \leq 1, \forall t$ and the $f(\bx_t; \btheta^{(0)}) \leq 1, \forall \bx_t$ by  lemma \ref{lemma:hbound} and choice of $m$.
\newline

\begin{lemma}\label{lemma:hbound}
Over the randomness of $\btheta_0$, with probability at least $1- O(tL) e^{- \Omega(m)}$, there exist constants $1< C_1, C_2 <2$, such that 
\[
\begin{aligned}
& \forall l \in {1, \dots, L}, \| h^{l, (0)}\|_F \leq  (C_1 \mu)^L, \ \ \|\bw^{l, (0)}\|_2 \leq C_1 \sqrt{qm},  \  \   \\
&f(\bx; \btheta^{(0)}) \leq \frac{(C_1 \mu)^L C_2}{\sqrt{m}}, \   \  \| \bw^{L+1, (0)} \|_2 \leq C_2 .
\end{aligned}
\]
\end{lemma}

\begin{lemma} \label{lemma:wbound}
In a round $t$, at $k$-th iteration of gradient descent, assuming $h^{l, (k)} \leq (C_1\mu)^L +  \Psi_{L, (k')} $  with probability at least
$1 - O(tL) e^{-\Omega(m)}$, we have

\[
\| \bw^{l, (k)} - \bw^{l, (0)} \|_F \leq \frac{w}{\sqrt{m}}, \ \text{for} \ 1\leq l \leq L+1,
\]
where 
\[
w = \frac{ 2 t \sqrt{2}  \mu^{L} e^{C_1(L-1)\sqrt{q} + C_2} ( (C_1 \mu)^L +  \Psi_{L, (k')})  } {C_0}.
\]
\end{lemma}

\begin{lemma} \label{lemma:hdiff}
In a round $t$, at $k$-th iteration of gradient descent, suppose $ \| \bw^{l, (k)} - \bw^{l, (0)} \|_F \leq w/\sqrt{m},  \forall l \in [T]$, with probability at least
$1 - O(tL) e^{-\Omega(m)}$, we have
\[
\begin{aligned}
&\forall l \in [L],  \| h^{l, (k)} - h^{l, (0)}\|_F \leq    \frac{\mu w \left( (2 \mu  C_1 \sqrt{q})^L -1 \right) }{m ( 2\mu  C_1 \sqrt{q} -1)} = \Psi_{L, (k')} \\
&\forall l \in [L],  \|  h^{l, (k)}  \|_F \leq (C_1\mu)^L +  \Psi_{L, (k')}.       
\end{aligned}
\]
\end{lemma}

\begin{definition}
For any layer $1\leq l \leq L-1$, we define
\[
\ch^{l} = m^{-\frac{1}{2}} qm^{-\frac{L-l+1}{2}}  {\bw^{L+1}}^\intercal \left( \prod_{j=l+1}^L \odot \bbd^{j}  (\bw^{j} \diamond \bbk) \right) \odot \bbd^{l}
\]
Therefore, $\triangledown_{\bw^l} f(\bx; \btheta)$ can be represented by
\[
\triangledown_{\bw^l} f(\bx; \btheta) =  \pphi{h^{l-1}} \ch^{l} .
\]
\end{definition}

\begin{lemma} \label{lemma:chbound}
In a round $t$, at $k$-th iteration, with probability at least
$1 - O(tL) e^{-\Omega(m)}$,  for $l \in [L]$, we have
\[
\begin{aligned} 
&\|\ch^{l, (0)}\|_F \leq   \frac{\mu \sqrt{p }}{m} (\mu C_1\sqrt{q})^{L-1} \\
&\|\ch^{l, (k)} - \ch^{l, (0)}\|_F \leq  \frac{\mu w \left( (2 \mu  C_1 \sqrt{q})^L -1 \right) }{m ( 2\mu  C_1 \sqrt{q} -1)} = \Psi_{L, (k')}\\
& \|\ch^{l, (k)}\|_F \leq   \frac{\mu \sqrt{p }}{m} (\mu C_1\sqrt{q})^{L-1} +  \Psi_{L, (k')}.
\end{aligned}
\]
\end{lemma}

\begin{lemma} \label{lemma:trif}
In a round $t$, at $k$-th, iteration, with probability at least $1 - O(tL) e^{-\Omega(m)}$,  for $l \in [L]$, we have
\[
\begin{aligned}
& \forall  1\leq l  \leq L+1 ,    \| \triangledown_{\bwl}f(\bx; \btheta^{(0)})\|_F \leq \sqrt{p}(C_1\mu\sqrt{q})^L/m,  \    \\
& \|\triangledown_{\bw^{L+1}} f(\bx; \btheta^{(k)})  -  \triangledown_{\bw^{L+1}}f(\bx; \btheta^{(0)}) \|_F \leq   \Psi_{L, (k')} \sqrt{q}
 \\
 & \forall l \in [L], \| \triangledown_{\bwl} f(\bx; \btheta^{(k)})  -  \triangledown_{\bwl}f(\bx; \btheta^{(0)}) \|_F \leq  \Psi_{L, (k')} \sqrt{q} ( (C_1\mu)^L + 2)
\end{aligned}
\]

\end{lemma}

\begin{lemma}\label{lemma:CNTKs}
In a round $t$, at $k$-th iteration of gradient descent, with probability at least
$1 - O(tL) e^{-\Omega(m)}$, we have
\[
\begin{aligned}
(1) \|g(\bx; \btheta^{(0)})\|_2 &\leq  \sqrt{(L+1)} \sqrt{p}(C_1\mu\sqrt{q})^L/m \\
(2) \| g(\bx_i; \btheta^{(k)}) - g(\bx_i; \btheta^{(0)})  \|_2 & \leq \sqrt{q(L+1})  \Psi_{L, (k')}  ( (C_1\mu)^L + 2) \\
(3) \|g(\bx_i; \btheta^{(k)})\|_2 & \leq  \sqrt{L+1} \left( \sqrt{p}(C_1\mu\sqrt{q})^L/m +   \sqrt{q} \Psi_{L, (k')}  ( (C_1\mu)^L + 2)   \right) 
\end{aligned}
\]
\end{lemma}

\begin{lemma}\label{lemma:f-g0}
In a round $t$, at $k$-th, iteration, with probability at least $1 - O(tL) e^{-\Omega(m)}$,  we have
\[
\begin{aligned}
&| f(\bx; \btheta^{(k)}) -  \dotp{g(\bx; \btheta^{(0)}),   \btheta^{(k)} - \btheta^{(0)} }|  \\
& \leq  \left( \Psi_{L, (k')} (C_2  + 1)  +  (C_1\mu)^L C_2 + \sqrt{q} w((L-1)  (C_1\mu)^L  + 1)  \right) /\sqrt{m}.
\end{aligned}
\]
\end{lemma}

\begin{lemma}\label{lemma:f-gk}
In a round $t$, at $k$-th, iteration, with probability at least $1 - O(tL) e^{-\Omega(m)}$,  we have
\[
\begin{aligned}
&| f(\bx; \btheta^{(k)}) -  \dotp{g(\bx; \btheta^{(k)}),   \btheta^{(k)} - \btheta^{(0)} }| \\
& \leq  \left \{ C_2 (\Psi_{L, (k')} + (C_1\mu)^L ) + \sqrt{q} (1 + \Psi_{L, (k')}  ) w  \left[(L-1) ( \Psi_{L, (k')} +  (C_1\mu)^L) + 1 \right] \right \}  /\sqrt{m}
\end{aligned}
\]
\end{lemma}

Lemma \ref{lemma:CNTKs} and \ref{lemma:f-g0} equals the Lemma $6.1$ and thus the proof is completed. Lemma \ref{lemma:f-gk} is the variant of Lemma $6.1$ used in the proof of Theorem $1$.

\subsection{Proofs}

\begin{prove}
\normalfont
According to \cite{vershynin2010introduction} and Lemma G.2 in \cite{du2019gradient},  with probability at least $1 - e^{-\frac{(c' - \sqrt{q} -1  )^2 m }{2}}$, there exists a constant $c'$,  for $\bw^l \in \bbr^{m \times qm}$, such that
\[
 \| \bw^l\|_2 \leq c' \sqrt{m}
\]
where $c'  > \sqrt{q} + 1$. Thus, we can derive
\[
 \| \bw^l \|_2 \leq  C_1 \sqrt{qm} , \ \forall l \in [L]
\]
with probability at least $1-O(L)e^{-\Omega(m)}$ and $ 1< C_1<2$.
For $\bw^{L+1} \in \bbr^{m \times p} $, applying Lemma in \cite{vershynin2010introduction} again, with probability at least $1 - e^{-\frac{(C_2 - \sqrt{p/m} - 1)^2m}{2}}$ we have 
\[
\sqrt{m} \| \bw^l \|_2 \leq  C_2\sqrt{m},
\]
where $1 <C_2 <2$ because $m > p$.

For $l \in [L]$, we have
\[
h^l = \frac{1}{\sqrt{qm}} \sigma( \bwl \pphi{h^{l-1}} ) \leq \frac{\mu}{\sqrt{qm}} \| \bwl\|_2 \| \pphi{h^{l-1}}\|_F  \leq  C_1 \mu \| h^{l-1} \|_F
\]
For $l=1$, we have
\[
h^1 \leq   \frac{\mu}{\sqrt{qm}} \| \bwl\|_2 \| \pphi{h^{l-1}} \|_F \leq C_1 \mu \| \bx\|_F \leq C_1 \mu
\]
Therefore, we have $h^l \leq (C_1\mu)^{L}$. Then apply the union bound with $L, t$.
\end{prove}
\qed

\begin{prove}
\normalfont

The induction hypothesis is $  \| \bw^{l, (k)} - \bw^{l, (0)} \|_F \leq  w/\sqrt{m}$.
To bound the gradient of one lay, we need the following claims:
\[
\begin{aligned}
&\prod_{j=l+1}^L \| \bbd^{j, (k)} \odot  (\bw^{j, (k)} \diamond \bbk)\|_2 \\
\leq & \prod_{j=l+1}^L \mu \|\bw^{j, (k)} \diamond \bbk\|_2 
\leq \prod_{j=l+1}^L \mu \sqrt{q} \|\bw^{j, (k)}\|_2 \\
\leq & \prod_{j=l+1}^L \mu \sqrt{q}\| ( \| \bw^{j, (0)} \|_2 + \|\bw^{j, (k)} - \bw^{j, (0)} \|_F) \\
\leq &  \mu^{L-l} q^{\frac{L-l}{2}} ( C_1 \sqrt{q} + w/m)^{L-l} m^{\frac{L-l}{2}}
\end{aligned}
\]
In $(k+1)$-th iteration, for $ 1 \leq l \leq L$, we have

\[
\begin{aligned}
&  \|\bw^{l, (k+1)} - \bw^{l, (k)} \|_F \\
= & m^{-\frac{1}{2}} (qm)^{-\frac{L-l+1}{2}}  \eta \sum_{i=1}^t |r_i - f(\bx_i; \btheta^{(k)})| \| {\phi(h^{l-1,(k)})}^{\intercal}  {\bw^{L+1, (k)}}^\intercal  \left( \prod_{j=l+1}^L \odot \bbd^{j, (k)} \bw^{j, (k) } \diamond \bbk \right) \odot \bbd^{l}       \|_F\\
\leq  & m^{-\frac{1}{2}} (qm)^{-\frac{L-l+1}{2}}\eta \sqrt{t} \|\mathbf{R}_t -\mathbf{F}_t^{(k)}\|_2 \|\bw^{L+1, (k)} \|_2  \mu^{L-l+1}  \prod_{j=l+1}^L \| \bw^{j, (k) } \diamond \bbk \|_2  \sqrt{q} \| h^{l-1, (k)}\|_F   \\
\leq & m^{-\frac{1}{2}} (qm)^{-\frac{L-l+1}{2}} \eta \sqrt{t} (1 - C_0 \eta)^{k/2} \| \mathbf{R}_t - \mathbf{F}_t^{(0)}\|_2 \| \bw^{L+1, (k)} \|_2  \\
&  \ \  \cdot   \mu^{L-l+1} q^{\frac{L-l}{2}} (C_1\sqrt{q} + w/m)^{L-l} m^{\frac{L-l}{2}} \cdot \sqrt{q} \| h^{l-1, (k)}   \|_F \\
\leq &   (1 - C_0 \eta)^{k/2} m^{-1} \eta \sqrt{t}  \| \mathbf{R}_t - \mathbf{F}_t^{(0)}\|_2 \mu^{L-l+1} e^{C_1(L-l)\sqrt{q} + C_2} \Psi_{h^{L, (k')}}  \\
\end{aligned}
\]
where the last inequality is because of Lemma \ref{lemma:hdiff} and $w/m \leq 1$.

For the layer $L+1$, we have

\[
\begin{aligned}
&  \|\bw^{L+1, (k+1)} - \bw^{L+1, (k)} \|_F \\
\leq &  \eta m^{-\frac{1}{2}}  \sum_{i=1}^t |r_i - f(\bx_i; \btheta^{(k)})| \| h^{L, (k)} \|_F  \\
\leq  &  \eta \sqrt{t}  m^{-\frac{1}{2}} \|\mathbf{R}_t -\mathbf{F}_t^{(k)}\|_2 \| h^{L, (k)}\|_F   \\
\leq &  (1 - C_0 \eta)^{k/2} m^{-\frac{1}{2}}  \eta \sqrt{t} \|\mathbf{R}_t -\mathbf{F}_t^{(0)}\|_2  \Psi_{h^{L, (k')}} 
\end{aligned}
\]
Thus,for $ 1 \leq l \leq L$, we have
\[
\begin{aligned}
&\| \bw^{l, (k+1)} - \bw^{l, (0)} \|_F  \leq \|\bw^{l, (k+1)} - \bw^{l, (k)}     \|_F +  \|\bw^{l, (k)} - \bw^{l, (0)}    \|_F\\
\leq & \sum_{i=0}^k    (1 - C_0 \eta)^{k/2} m^{-1}  \eta \sqrt{t}  \| \mathbf{R}_t - \mathbf{F}_t^{0}\|_2  \mu^{L-l+1} e^{C_1(L-l)\sqrt{q} + C_2}  \Psi_{h^{L, (k')}} \\
\leq & \frac{2}{C_0} m^{-1}  \sqrt{t}  \| \mathbf{R}_t - \mathbf{F}_t^{0}\|_2 \mu^{L} e^{C_1(L-1)\sqrt{q} + C_2}  \Psi_{h^{L, (k')}} \leq \frac{w}{\sqrt{m}}.
\end{aligned}
\]
For  the layer $L+1$, we have, 
\[
\begin{aligned}
&\| \bw^{L+1, (k+1)} - \bw^{L+1, (0)} \|_F  \\
\leq & \|\bw^{L+1, (k+1)} - \bw^{L+1, (k)}     \|_F +  \|\bw^{L+1, (k)} - \bw^{L+1, (0)}    \|_F\\
\leq &  \sum_{i=0}^k (1 - C_0 \eta)^{k/2}  m^{-\frac{1}{2}}  \eta \sqrt{t} \|\mathbf{R}_t -\mathbf{F}_t^{(0)}\|_2   \Psi_{h^{L, (k')}} \\
\leq &  \frac{2}{C_0}  m^{-\frac{1}{2}} \sqrt{t}  \| \mathbf{R}_t - \mathbf{F}_t^{0}\|_2  \Psi_{h^{L, (k')}} \leq   \frac{w}{\sqrt{m}}.
\end{aligned}
\]

The proof is completed.
\end{prove}
\qed

\begin{prove}
\normalfont

We prove this lemma by induction. The induction hypothesis is $\| h^{l, (k)} - h^{l, (0)}\|_F \leq \frac{\mu w}{m} g(l)$. 
\[
\begin{aligned}
&\| h^{l, (k)} - h^{l, (0)}\|_F  = \frac{1}{\sqrt{qm}} \left \| \bbd^{l}\odot \left ( \bw^{l, (k)}  \pphi{h^{l-1, (k)}} \right)     - \bbd^{l}\odot \left ( \bw^{l, (0)} \pphi{h^{l-1, (0)}} \right)      \right \|_F \\
\leq & \frac{1}{\sqrt{qm}}  \mu \left( \|\bw^{l, (k)}  \pphi{h^{l-1, (k)}}  -  \bw^{l, (k)}  \pphi{h^{l-1, (0)}}\|_F + \|\bw^{l, (k)}  \pphi{h^{l-1, (0)}}  -  \bw^{l, (0)}  \pphi{h^{l-1, (0)}}\|_F  \right)\\
= & \frac{1}{\sqrt{qm}}  \|   \bw^{l, (k)}  \left(   \pphi{h^{l-1, (k)}} - \pphi{h^{l-1, (0)}}   \right) \|_F
 + \frac{1}{\sqrt{qm}} \mu \|  \left(\bw^{l, (k)} - \bw^{l, (0)}    \right)    \pphi{h^{l-1, (0)}}  \|_F \\
\leq & \frac{1}{\sqrt{m}} \mu \left( \|\bw^{l, (0)}\|_2  + \|\bw^{l, (k)}- \bw^{l, (0)}\|_F  \right)  \cdot\|h^{l-1, (k)} - h^{l-1, (0)}\|_F  \\
& +  \frac{1}{\sqrt{m}}\mu  \| h^{l-1, (0)} \|_F \| \bw^{l , (k)} -  \bw^{l, (0)} \|_F \\
\leq & \mu (C_1\sqrt{q} + w/m) \|h^{l-1, (k)} - h^{l-1, (0)}\|_F + \mu w/m \\
\leq & \mu (C_1 \sqrt{q} + w/m) \mu \frac{w}{m} g(l-1) + \mu w/m\\
 = &  \mu \frac{w}{m} (\mu C_1 \sqrt{q} g(l-1) + \mu \frac{w}{m}  g(l-1) + 1)\\
 \leq &  \mu \frac{w}{m} (2\mu C_1 \sqrt{q} g(l-1) + 1)\\
=   &  \mu \frac{w}{m} g(l) \\
\end{aligned}
\]

Then, for $l =1$,  we have
\[
\begin{aligned}
&\| h^{1, (k)} - h^{1, (0)}\|_F  = \frac{\mu}{\sqrt{qm}} \left \| \bw^{1, (k)}  \pphi{\bx}    -  \bw^{1, (0)} \pphi{\bx}  \right \|_F \\
\leq & \frac{\mu}{\sqrt{m}} ( \|  \bw^{1, (k)} -   \bw^{1, (0)}   \|_F \| \bx \|_F  )\\
\leq & \frac{\mu w }{m}   =  \frac{\mu w }{m} g(1),
\end{aligned}
\]
where $g(1) = 1$. By calculation, we have
\[
\begin{aligned}
&\forall l \in [L],  \| h^{l, (k)} - h^{l, (0)}\|_F \leq   \frac{\mu w \left( (2 \mu  C_1 \sqrt{q})^L -1 \right) }{m ( 2\mu  C_1 \sqrt{q} -1)} = \Psi_{L, (k')} \\
&\forall l \in [L],  \|  h^{l, (k)}  \|_F \leq \| h^{l, (0)}\|_F +   \| h^{l, (k)} - h^{l, (0)}\|_F \leq \mu^L +  \Psi_{h^{L, (k')}}.      
\end{aligned}
\]
The proof is completed.
\end{prove}

\qed

\begin{prove}
\normalfont

Define ${\bbd^{l}}' = \bbd^{l} - \bbd^{l} $ To show the results, we need the following claims.
\[
\begin{aligned}
&\| \bw^{l+1, (k)} \diamond \bbk \odot \bbd^{l}  -  \bw^{l+1, (0)} \diamond \bbk \odot \bbd^{l}  \|_F\\
\leq & \mu   \| \bw^{l+1, (k)} \diamond \bbk   -  \bw^{l+1, (0)} \diamond \bbk \|_F\\
\leq & \mu \sqrt{q} \| \bw^{l+1, (k)} -  \bw^{l+1, (0)}  \|_F\\
\leq & \frac{\mu w}{m} \sqrt{qm}
\end{aligned}
\]

Then, we have
\[
\begin{aligned}
& \|\ch^{l, (k)} - \ch^{l, (0)}\|_F  \\
= & m^{-\frac{1}{2}}  (qm)^{-\frac{L-l+1}{2}}  \|    \bw^{L+1, (k)} \left( \prod_{j=l+1}^L \cdot \bbd^{j,(k)} \bw^{j, (k)} \diamond \bbk \right) \bbd^{l} \\
&-  \bw^{L+1, (0)} \left( \prod_{j=l+1}^L \odot \bbd^{j,(k)} \bw^{j, (0)}  \diamond \bbk \right)  \bbd^{l} \|_F \\
\leq &   \frac{1}{\sqrt{qm}}\| \ch^{l+1, (k)} \bw^{l+1, (k)} \diamond \bbk \odot \bbd^{l}  
-  \ch^{l+1, (0)}  \bw^{l+1, (k)} \diamond \bbk  \odot \bbd^{l} \|_F  \\   
& +  \frac{1}{\sqrt{qm}}  \| \ch^{l+1, (0)}   \bw^{l+1, (k)} \diamond \bbk \odot \bbd^{l} - \ch^{l+1, (0)}  \bw^{l+1, (0)} \diamond \bbk  \odot \bbd^{l} \|_F \\
\leq &  \frac{\mu}{\sqrt{qm}}  \| \ch^{l+1, (k)} - \ch^{l+1, (0)} \|_F \|   \bw^{l+1, (k)} \diamond \bbk \|_2 \\
&   +  \frac{1}{\sqrt{qm}} \| \ch^{l+1, (0)}\|_F \| \bw^{j, (k)} \diamond \bbk \odot \bbd^{l}  -  \bw^{j, (0)}  \diamond \bbk \odot \bbd^{l}  \|_2 \\ 
\leq &  \mu (C_1\sqrt{q}+w/m) \| \ch^{l+1, (k)} - \ch^{l+1, (0)} \|_F   \  +  \mu w/m \\
\leq &  \mu (C_1\sqrt{q}+w/m) \frac{\mu w}{m} g(l+1) \  +  \mu w/m  \\
\leq & \frac{\mu w}{m} (2 \mu C_1\sqrt{q}   g(l+1) +1) \\
\leq & \frac{\mu w}{m} g(l) 
\end{aligned}
\]

For $L$-th layer, we have 

\[
\begin{aligned}
&\|\ch^{L, (k)} - \ch^{L, (0)}\|_F \\
\leq &  m^{-\frac{1}{2}}     \frac{1}{\sqrt{qm}} \| \bw^{L+1, (k)} \odot \bbd^{l} - \bw^{L+1, (0)} \odot \bbd^{l}\|_F \\
\leq &  m^{-\frac{1}{2}}   \frac{ \mu}{\sqrt{qm}} \| \bw^{L+1, (k)} - \bw^{L+1, (0)}\|_F \\
\leq &    \frac{\mu w}{m}  \frac{1}{\sqrt{qm}}   \leq  \frac{\mu w}{m} g(L)
\end{aligned}
\]
where $g(L) = 1$.   
Therefore, by calculation, we have 
\[
\|\ch^{l, (k)} - \ch^{l, (0)}\|_F \leq  
\frac{\mu w \left( (2 \mu  C_1 \sqrt{q})^L -1 \right) }{m ( 2\mu  C_1 \sqrt{q} -1)} = \Psi_{L, (k')}
\]

For $\ch^{l,(0)}$, we have 
\[
\begin{aligned}
&\|  \ch^{l, (0)} \|_F \leq m^{-\frac{1}{2}} qm^{-\frac{L-l+1}{2}} \|\bw^{L+1, (0)}  \left( \prod_{j=l+1}^L \odot \bbd^{j,(0)} \bw^{j, (0)} \diamond \bbk \right) \odot \bbd^{l} \|_F  \\
\leq &   m^{-\frac{1}{2}} qm^{-\frac{L-l+1}{2}} \mu  \|  \bw^{L+1, (0)} \left( \prod_{j=l+2}^L \odot \bbd^{j,(0)} \bw^{j, (0)} \diamond \bbk \right) \odot \bbd^{l+1, (0)}\|_F \| \bw^{l+1, (0)} \diamond \bbk\|_2 \\
\leq & \mu C_1\sqrt{q} \| \ch^{l+1, (0)}\|_F 
\end{aligned}
\]
For the lay $L$, we have
\[
\|\ch^{L, (0)}\|_F = \frac{1}{\sqrt{q}m} \| \bw^{L+1, (0)} \odot \bbd^{L}\|_F \leq \frac{\mu \sqrt{p }}{\sqrt{q}m}.
\]
By caculation, we have
\[
\|\ch^{l, (0)}\|_F \leq  \frac{\mu \sqrt{p }}{m} (\mu C_1\sqrt{q})^{L-1} \leq 1 .
\]
The proof is completed.
\end{prove}
\qed

\begin{prove}
\normalfont
By Lemma \ref{lemma:hbound}, \ref{lemma:hdiff}, \ref{lemma:chbound}, for $l \in [L]$, we have
\[
\| \triangledown_{\bwl}f(\bx; \btheta^{(0)})\|_F = \| \ch^{l, (0)}  \pphi{h^{l-1, (0)}} \|_F \leq  \sqrt{q} \| \ch^{l, (0)}    \|_F    \| h^{l-1, (0)}\|_F \leq    \sqrt{p}(C_1\mu\sqrt{q})^L/m.
 \]
For $L+1$ layer, we have $ \| \triangledown_{\bw^{L+1, (0)}}f(\bx; \btheta^{(0)})\|_F = \| h^{L, (0)}\|_F \leq 1$.

\[
\begin{aligned}
& \| \triangledown_{\bwl} f(\bx; \btheta^{(k)})  -  \triangledown_{\bwl}f(\bx; \btheta^{(0)}) \|_F \\
= & \|\ch^{l, (k)}  \pphi{h^{l-1, (k)}}  - \ch^{l, (0)}  \pphi{h^{l-1, (0)}}  \|_F \\ 
\leq & \left \| \ch^{l, (k)}    \pphi{h^{l-1, (k)}}  - \ch^{l, (0)}  \pphi{h^{l-1,(k)}} \right \|_F  + \left \|  \ch^{l, (0)}  \pphi{h^{l-1,(k)}} -  \ch^{l, (0)}  \pphi{h^{l-1, (0)}}   \right \|_F \\
\leq & \left\|  \ch^{l, (k)} -  \ch^{l, (0)}\right\|_F \left( \left \|    \pphi{(h^{l-1, (0)}} \right \|_F +  \left\| \pphi{(h^{l-1, (k)}} - \pphi{(h^{l-1, (0)}}  \right\|_F \right)   \\
& + \left \|\ch^{l, (0)}  \right\|_F  \left \| \pphi{(h^{l-1, (k)}} - \pphi{(h^{l-1, (0)}} \right \|_F\\
\leq &  \sqrt{q} \Psi_{L, (k')} ( (C_1\mu)^L +  \Psi_{L, (k')}) +   \Psi_{L, (k')} \\
\leq & \Psi_{L, (k')} \sqrt{q} ( (C_1\mu)^L + 2)
\end{aligned}
\]

For $L+1$ layer, we have $\|\triangledown_{\bw^{L+1}} f(\bx; \btheta^{(k)})  -  \triangledown_{\bw^{L+1}}f(\bx; \btheta^{(0)}) \|_F =  \|\pphi{h^{L, (k)}} - \pphi{h^{L, (0)}}\|_F  \leq \Psi_{L, (k')} \sqrt{q} $.

This proof is completed.
\end{prove}
\qed

\begin{prove}

For (1), we have
\[
\|g(\bx; \btheta^{(0)})\|_2  = \sqrt{\sum_{l=1}^{L+1} \|\triangledown_{\bwl} f(\bx; \btheta^{(0)})\|_F^2 } 
\leq  \sqrt{(L+1)} \sqrt{p}(C_1\mu\sqrt{q})^L/m \\
\]

For (2), we have
\[
\| g(\bx_i; \btheta^{(k)}) - g(\bx_i; \btheta^{(0)})  \|_2 
= \sqrt{ \sum_{l=1}^{L+1} \| \triangledown_{\bwl} f(\bx; \btheta^{(k)}) -  \triangledown_{\bwl} f(\bx; \btheta^{(0)})     \|_F^2}  \leq \sqrt{q(L+1})  \Psi_{L, (k')}  ( (C_1\mu)^L + 2).
\]

For (3), we have
\[
\|g(\bx_i; \btheta^{(k)})\|_2 =  
 \sqrt{\sum_{l=1}^{L+1} \|\triangledown_{\bwl} f(\bx; \btheta^{(k)})\|_F^2 }  \leq   \sqrt{L+1} \left( \sqrt{p}(C_1\mu\sqrt{q})^L/m +   \sqrt{q} \Psi_{L, (k')}  ( (C_1\mu)^L + 2)   \right) 
\]
With Lemma \ref{lemma:trif}, the proof is completed.
\end{prove}
\qed

\begin{prove}
\normalfont

By Lemma \ref{lemma:hbound}, \ref{lemma:wbound}, \ref{lemma:hdiff}, \ref{lemma:chbound}, we have  

\[
\begin{aligned}
&| f(\bx; \btheta^{(k)}) -  \dotp{g(\bx; \btheta^{(0)}),   \btheta^{(k)} - \btheta^{(0)} }|\\
= & |\langle  h^{L, (k)}  ,\bw^{L+1, (k)}\rangle / \sqrt{m} -   \langle h^{L, (0)},  (\bw^{L+1, (k)} -  \bw^{L+1, (0)})    \rangle / \sqrt{m}  \\
&-  \sum_{l=1}^L \ch^{l, (0)}  (\bw^{l,(k)} - \bw^{l, (0)})  \pphi{h^{l-1, (0)} }   |\\
= &\left |\langle  h^{L, (k)} - h^{L, (0)},\bw^{L+1, (k)}\rangle / \sqrt{m} +  \langle h^{L, (0)},  \bw^{L+1, (0)}\rangle /\sqrt{m}  - \sum_{l=1}^L \ch^{l, (0)}  (\bw^{l,(k)} - \bw^{l, (0)})  \pphi{h^{l-1, (0)} } \right | \\
\leq  &  \|  h^{L, (k)} - h^{L, (0)}  \|_F \| \bw^{L+1, (k)}\|_2 / \sqrt{m}+ \|h^{L, (0)}\|_F \| \bw^{L+1,(0)}  \|_2 / \sqrt{m}  \\
& +  \sum_{l=1}^L \|\ch^{l, (0)}\|_F  \|\bw^{l,(k)} - \bw^{l, (0)}  \|_F  \| \pphi{h^{l-1, (0)}}\|_F \\
\leq &  \Psi_{L, (k')} (C_2  + 1) / \sqrt{m} +  (  (C_1\mu)^L C_2)/\sqrt{m} + \sqrt{q} w((L-1)  (C_1\mu)^L  + 1) / \sqrt{m} \\
=  & \left( \Psi_{L, (k')} (C_2  + 1)  +  (C_1\mu)^L C_2 + \sqrt{q} w((L-1)  (C_1\mu)^L  + 1)  \right) /\sqrt{m}.
\end{aligned}
\]
The proof is completed.
\end{prove}
\qed

\begin{prove}
\normalfont

By Lemma \ref{lemma:hbound}, \ref{lemma:wbound}, \ref{lemma:hdiff}, \ref{lemma:chbound}, we have  

\[
\begin{aligned}
&| f(\bx; \btheta^{(k)}) -  \dotp{g(\bx; \btheta^{(k)}),   \btheta^{(k)} - \btheta^{(0)} }|\\
= & |\langle  h^{L+1, (k)}  ,\bw^{L+1, (k)}\rangle / \sqrt{m} -   \langle h^{L+1, (k)},  (\bw^{L+1, (k)} -  \bw^{L+,1 (0)})    \rangle /\sqrt{m}  \\
&+  \sum_{l=1}^L \ch^{l, (k)}  (\bw^{l,(k)} - \bw^{l, (0)})  \pphi{h^{l-1, (k)} }   |\\
= & \left | \langle \bw^{L+1,(0)}, h^{L, (k)} \rangle/\sqrt{m} - \sum_{l=1}^{L} \ch^{l, (k)}  (\bw^{l,(k)} - \bw^{l, (0)})  \pphi{h^{l-1, (k)} }    \right |\\
\leq &  \| \bw^{L+1, (0)}\|_2 \| h^{L, (k)}  \|_F /\sqrt{m}  +  \sum_{l=1}^{L} \left(\| \ch^{l, (k)}\|_F \right) \| \bw^{l,(k)} - \bw^{l, (0)}   \|_F \sqrt{q} \| h^{l-1, (k)}\|_F \\
\leq & C_2 (\Psi_{L, (k')} + (C_1\mu)^L) / \sqrt{m} + \sqrt{q} (1 + \Psi_{L, (k')}  ) w  \left[(L-1) ( \Psi_{L, (k')} + (C_1\mu)^L) + 1 \right] /\sqrt{m}  \\
= &  \left \{ C_2 (\Psi_{L, (k')} + (C_1\mu)^L ) + \sqrt{q} (1 + \Psi_{L, (k')}  ) w  \left[(L-1) ( \Psi_{L, (k')} +  (C_1\mu)^L) + 1 \right] \right \}  /\sqrt{m}.
\end{aligned}
\]
The proof is completed.
\end{prove}
\qed

\section{Proof of Lemma 6.2}

\begin{definition} \label{def:ridge}
Given the context vectors $\{\bx_i\}_{i=1}^t$ and the rewards $\{ r_i \}_{i=1}^{t} $, then we define the estimation $\widehat{\btheta}_0$ via ridge regression:  
\[
\begin{aligned}
&\mathbf{A}_0 =  \lambda \mathbf{I} + \sum_{i=1}^{t} g(\bx_i; \btheta^{(0)}) g(\bx_i; \btheta^{(0)})^\intercal /m \\
&\mathbf{b}_0 = \sum_{i=1}^t r_t g(\bx_i; \btheta^{(0)})  /\sqrt{m}\\
&\widehat{\btheta}_0 = \mathbf{A}^{-1}_0\mathbf{b}_0 
\end{aligned}
\]
\end{definition}

\begin{definition}
\[
\begin{aligned}
&\mathbf{G}^{(k)} = \left( g(\bx_1; \btheta^{(k)}), \dots,   g(\bx_t; \btheta^{(k)})\right)\\
&\mathbf{G}^{(0)} = \left( g(\bx_1; \btheta^{(0)}), \dots,   g(\bx_t; \btheta^{(0)})\right)\\
&\mathbf{f}^{(k)} = \left( f(\bx_1; \btheta^{(k)}), \dots,   f(\bx_t; \btheta^{(k)})    \right)^\intercal \\
& \mathbf{r} = \left( r_1, \dots, r_t  \right)^\intercal \\
& \btheta^{(k+1)}  = \btheta^{(k)} -\eta \left[ \bsg^{(k)}(\bsf^{(k)} - \bsr )\right]\\
\end{aligned}
\]
Inspired by Lemma B.2 in \cite{zhou2020neural} , we define the auxiliary sequence following :
\[
\wbtheta^{(0)} = \btheta^{(0)}, \ \ \wbtheta^{(k+1)}   = \wbtheta^{(k)} - \eta\left[ \bsg^{(k)} \left( [\bsg^{(k)}]^\intercal (\wbtheta^{(k)}  - \wbtheta^{(0)}) - \bsr \right)  +  m\lambda (\wbtheta^{(k)} - \wbtheta^{(0)} )  \right] 
\]
\end{definition}

\begin{lemma}\label{lemma:subg}
In a round $t$, at $k$-th iteration of gradient descent, with probability at least
$1 - O(pL) e^{-\Omega(m)}$, we have
\[
\begin{aligned}
(1) \|\bsg^{(0)}\|_F & \leq \sqrt{t(L+1)} \sqrt{p}(C_1\mu\sqrt{q})^L/m \\
(2) \|\bsg^{(k)} - \bsg^{(0)}\|_F & \leq \sqrt{tq(L+1})  \Psi_{L, (k')}  ( (C_1\mu)^L + 2) \\
(3) \|\bsg^{(k)}\|_F  & \leq \sqrt{t(L+1)} \sqrt{p}(C_1\mu\sqrt{q})^L/m  + \sqrt{tq(L+1})  \Psi_{L, (k')}  ( (C_1\mu)^L + 2) =   \bar{A}_1 \\
(4)  \|\mathbf{f}^{k} - {\bsg^{(k)}}^\intercal ( \btheta^{(k)} - \btheta^{(0)} )\|_2 & \leq \sqrt{t} \left \{ C_2 (\Psi_{L, (k')} + (C_1\mu)^L ) + \sqrt{q} (1 + \Psi_{L, (k')}  ) w  \left[(L-1) ( \Psi_{L, (k')} +  (C_1\mu)^L) + 1 \right] \right \}  /\sqrt{m}\\
& = \sqrt{t} \Psi_3  .
\end{aligned}
\]
\end{lemma}

\begin{lemma} \label{lemma:2thetab}
In a round $t$, at $k$-th iteration of gradient descent, with probability at least
$1 - O(pL) e^{-\Omega(m)}$, we have
\[
\begin{aligned}
& (1) \| \wbtheta^{(k)} - \btheta^{(0)} - \widehat{\btheta}_t/\sqrt{m} \|_2  \leq \sqrt{\frac{t}{m \lambda}}  \\
\end{aligned}
\]
\end{lemma}

\noindent \textbf{Proof of Lemma 6.2}
\[
\begin{aligned}
&\| \wbtheta^{(k+1)} - \btheta^{(k+1)} \|_2  \\
= & \|  \wbtheta^{(k)} - \btheta^{(k)}  - \eta\left[ \bsg^{(k)} \left( [\bsg^{(k)}]^\intercal (\wbtheta^{(k)}   - \btheta^{(k)} +  \btheta^{(k)} - \wbtheta^{(0)}) - \bsr \right)  +  m\lambda (\wbtheta^{(k)} - \btheta^{(k)} +  \btheta^{(k)} - \wbtheta^{(0)} ) \right]   \\
& +  \eta \left[ \bsg^{(k)}(\bsf^{(k)} - \bsr )\right]    \|  \\
=&  \| (1-\eta m\lambda) (\btheta^{(k)} -  \wbtheta^{(k)}) 
- \eta \bsg^{(k)} \left(  [\bsg^{k}]^{\intercal} (\btheta^{(k)} - \btheta^{(0)}) - \bsf^{(k)}   \right) \\
& - \eta \bsg^{(k)}{\bsg^{(k)}}^\intercal  (\btheta^{(k)} -  \wbtheta^{(k)})    - \eta m\lambda   (\btheta^{(k)} - \btheta^{(0)})  \|_2 \\
\leq & 
 \eta \| \bsg^{(k)} \|_2\|\bsf^{(k)} -  [\bsg^{k}]^{\intercal} (\btheta^{(k)} - \btheta^{(0)})   \|_2  \\
&+ \eta m \lambda \| \btheta^{(k)} - \btheta^{(0)}\|_2 + \left\| \mathbf{I} - \eta(m\lambda \mathbf{I} +  \bsg^{(k)}{\bsg^{(k)}}^\intercal )   \right\|_2 \| \btheta^{(k)} - \wbtheta^{(k)}\|_2 \\
=& A_1 + A_2 + A_3 
\end{aligned}
\] 
For $A_1$, by Lemma \ref{lemma:subg}, we have
\[
A_2 \leq  \eta \cdot \bar{A}_1 \cdot \sqrt{t} \Psi_3.
\]
For $A_2$, by Lemma \ref{lemma:wbound}, we have
\[
\eta  m \lambda \| \btheta^{(k)} - \btheta^{(0)}\|_2 = \eta m \lambda  \sqrt{\sum_{l=1}^{L+1} \|\bw^{l, (k)} - \bw^{l, (0)}\|_F^2} =  \eta  m \lambda \sqrt{L+1}w/ \sqrt{m} = \eta m \bar{A}_2.
\]
For $A_3$, by Lemma \ref{lemma:subg}, we have 
\[
A_3  \leq  (1 - \eta m \lambda) \| \btheta^{(k)} - \wbtheta^{(k)}\|_2   
\]
because $\mathbf{I} - \eta(m\lambda \mathbf{I} +  \bsg^{(k)}{\bsg^{(k)}}^\intercal ) \preceq  (1 - \eta m \lambda)\mathbf{I}$ with the choice of $m$ and $\eta$.
Therefore, adding everything together, we have 
\begin{equation}
\| \btheta^{(k+1)} - \wbtheta^{(k+1)}\|_2 \leq  (1 - \eta m \lambda) \| \btheta^{(k)} - \wbtheta^{(k)}\|_2  +    \eta \cdot \bar{A}_1 \cdot \sqrt{t} \Psi_3 +  \eta m \bar{A}_2
\end{equation}
As $\|\btheta^{(0)} - \wbtheta^{(0)}\| = 0$, by induction, we have
\begin{equation}
\| \btheta^{(k)} - \wbtheta^{(k)}\|_2 \leq \frac{  \eta \cdot \bar{A}_1 \cdot \sqrt{t} \Psi_3 +  \eta m \bar{A}_2 }{ \eta m \lambda } \leq \frac{    \bar{A}_1 \cdot \sqrt{t} \Psi_3 +   m \bar{A}_2 }{ m \lambda }
\end{equation}
By Lemma \ref{lemma:2thetab}, we have
\[
\begin{aligned}
\| \btheta^{(k)} - \btheta^{(0)} -  \widehat{\btheta}_t /\sqrt{m}\|_2 = & \| \btheta^{(k)} - \wbtheta^{(k)} + \wbtheta^{(k)} - \btheta^{(0)}  - \widehat{\btheta}_t /\sqrt{m}\|_2\\
\leq &   \| \btheta^{(k)} - \wbtheta^{(k)}     \|_2 +  \| \wbtheta^{(k)} - \btheta^{(0)} - \widehat{\btheta}_t/\sqrt{m}  \|_2 \\
\leq &\frac{    \bar{A}_1 \cdot \sqrt{t} \Psi_3 +   m \bar{A}_2 }{ m \lambda } + \sqrt{  \frac{t}{m\lambda} }
\end{aligned}
\]
The proof is completed.
\qed
\newline

\begin{prove}
\normalfont
By Lemma \ref{lemma:trif}, \ref{lemma:f-g0}, we have
For (1), we have
\[
\|\bsg^{(0)}\|_F \leq \sqrt{\sum_{i=1}^{t} \| g(\bx_1; \btheta^{(0)})  \|_2^2 }  \leq \sqrt{t(L+1)} \sqrt{p}(C_1\mu\sqrt{q})^L/m. 
\]
For (2), we have
\[
\begin{aligned}
&  \|\bsg^{(k)} - \bsg^{(0)}\|_F = \sqrt{\sum_{i=1}^t \|g(\bx_i; \btheta^{(k)}) - g(\bx_i; \btheta^{(0)})\|^2_2}  \\
&\leq \sqrt{tq(L+1})  \Psi_{L, (k')}  ( (C_1\mu)^L + 2)  \\
\end{aligned}
\]
For (3), we have
\[
\| \bsg^{(k)}\|_F \leq  \|\bsg^{(k)} - \bsg^{(0)}\|_F + \|\bsg^{(0)}\|_F \leq \bar{A}_1.
\]
For (4), based on Lemma \ref{lemma:f-gk},  we have
\[
\begin{aligned}
&\|\mathbf{f}^{k} - {\bsg^{(k)}}^\intercal ( \btheta^{(k)} - \btheta^{(0)} )\|2 = \sqrt{\sum_{i=1}^t | f(\bx_i; \btheta^{(k)}) -  \dotp{g(\bx_i; \btheta^{(k)}),   \btheta^{(k)} - \btheta^{(0)} }|^2} \\
&\leq \sqrt{t} \left \{ C_2 (\Psi_{L, (k')} + (C_1\mu)^L ) + \sqrt{q} (1 + \Psi_{L, (k')}  ) w  \left[(L-1) ( \Psi_{L, (k')} +  (C_1\mu)^L) + 1 \right] \right \}  /\sqrt{m}.
\end{aligned}
\]
\end{prove}
\qed

\begin{prove}
\normalfont
The sequence of $\wbtheta^{(k)}$ is updates by using gradient descent on the loss function:
\[
\min_{\wbtheta} \mathcal{L}(\wbtheta) = \frac{1}{2}  \|[\bsg^{(k)}]^\intercal (\wbtheta - \btheta^{(0)} ) - \bsr  \|^2_2 + \frac{m\lambda}{2} \|\wbtheta - \btheta^{(0)}\|_2^2 . 
\]

By standard results of gradient descent on ridge regression, $\wbtheta^{(k)}$ converges to $\btheta^{(0)} + \widehat{\btheta}_t /\sqrt{m} $. Therefore, we have 
\[
\begin{aligned}
\| \wbtheta^{(k)} - \btheta^{(0)} - \widehat{\btheta}_t / \sqrt{m} \|_2^2  & \leq \left[  1 -\eta m\lambda\right]^k \frac{2}{m\lambda} \left(  \mathcal{L}(\btheta^{(0)}) - L(\btheta^{(0)} + \widehat{\btheta}_t / \sqrt{m}) \right) \\
 \leq &  \frac{2 (1 - \eta m \lambda)^k}{m\lambda}  \mathcal{L}(\btheta^{(0)}) \\
= & \frac{2 (1 - \eta m \lambda)^k}{ m \lambda}  \frac{\|\bsr\|^2}{2} \\
\leq &\frac{t(1 - \eta m\lambda)^k}{m\lambda}  \leq  \frac{t}{m \lambda}.
\end{aligned}
\]
\end{prove}
\qed

\end{document}